\documentclass[10pt,twocolumn,letterpaper]{article}

\usepackage{iccv}
\usepackage{times}
\usepackage{epsfig}
\usepackage{graphicx}
\usepackage{amsmath}
\usepackage{amssymb}
\usepackage{color}
\usepackage{subfigure}
\usepackage{multirow}

\graphicspath{{figures_jpg/}}

\def\degree{${}^{\circ}$}

\usepackage[breaklinks=true,bookmarks=false]{hyperref}

\iccvfinalcopy 

\def\iccvPaperID{****} 

\begin{document}

\title{Kernelized Deep Convolutional Neural Network for Describing Complex Images}

\author{Zhen Liu\\
University of Science and Technology of China\\
{\tt\small liuzheng@mail.ustc.edu.cn}
}

\maketitle

\begin{abstract}
   With the impressive capability to capture visual content, deep convolutional neural networks (CNN) have demonstrated promising performance in various vision-based applications, such as classification, recognition, and object detection. However, due to the intrinsic structure design of CNN, for images with complex content, it achieves limited capability on invariance to translation, rotation, and re-sizing changes, which is strongly emphasized in the scenario of content-based image retrieval. In this paper, to address this problem, we proposed a new kernelized deep convolutional neural network. We first discuss our motivation by an experimental study to demonstrate the sensitivity of the global CNN feature to the basic geometric transformations. Then, we propose to represent visual content with approximate invariance to the above geometric transformations from a kernelized perspective.  We extract CNN features on
   the detected object-like patches and aggregate these patch-level CNN features to form a vectorial representation with
   the Fisher vector model. The effectiveness of our proposed algorithm is demonstrated on image search application with three benchmark datasets.
\end{abstract}

\section{Introduction}

Vectorial image representation is a fundamental problem in computer vision field. In many visual analysis systems, the visual content in an image is usually represented into a fix-sized vector for convenience of the followed processing. In recent years a lot of effort has been made on first designing the handcraft visual features \cite{SIFT, HOG, bay2006surf} and then aggregating the visual features into a single vector \cite{video:google, wang2010locality, FisherVector, jegou2010aggregating, jegou:hal-00977321}.

The bag-of-visual-words (BoVW) model is one of the famous methods to construct image representation. In the BoVW model, firstly, a set of local invariant visual features are extracted on the detected image patches or the densely sampled grids. Then an image is represented into a visual word histogram based on the quantization results of local features with an off-line trained visual vocabulary. The visual vocabulary is usually trained with the unsupervised clustering algorithm, such as the standard $k$-means, hierarchical $k$-means \cite{vocabuary:tree}, approximate $k$-means \cite{AKM}. Usually the quantization is performed by the nearest neighbor or the approximate nearest neighbor method. Namely each local invariant visual feature is quantized to its nearest or approximate nearest visual word in the vocabulary, which is a kind of hard vector quantization. Instead of the hard vector quantization, in \cite{wang2010locality}, Wang et al. proposed a locality linear coding approach to quantize each local visual feature.

\begin{figure*}[t]
\begin{center}
 \subfigure[]{
 \includegraphics[width=0.4 \linewidth]{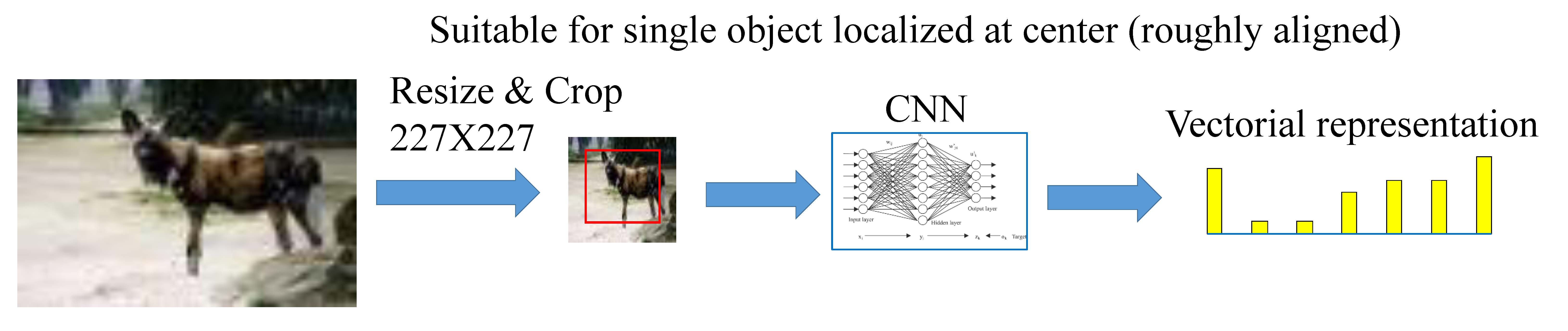}
 \label{cnn_s1}
 }
 \subfigure[]{
 \includegraphics[width=0.4 \linewidth]{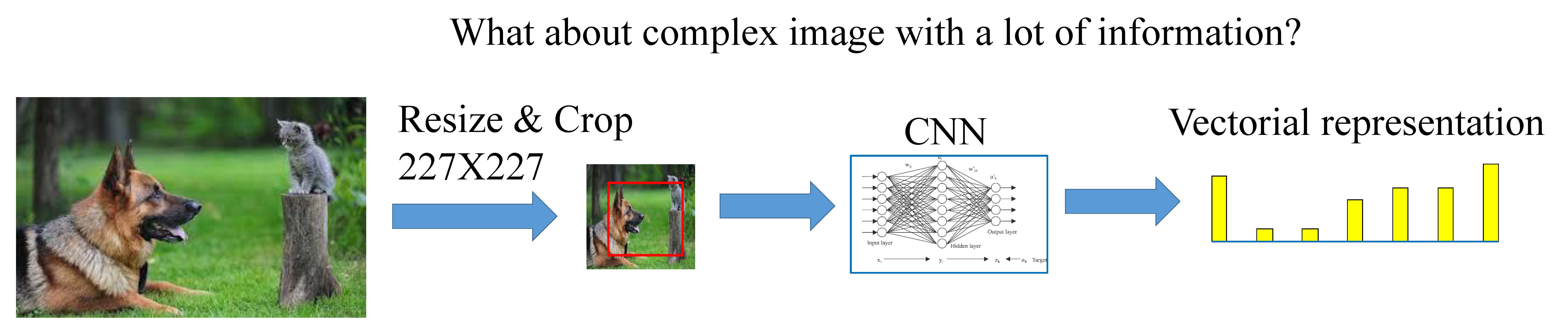}
  \label{cnn_s2}
 }
 \subfigure[]{
 \includegraphics[width=0.8 \linewidth]{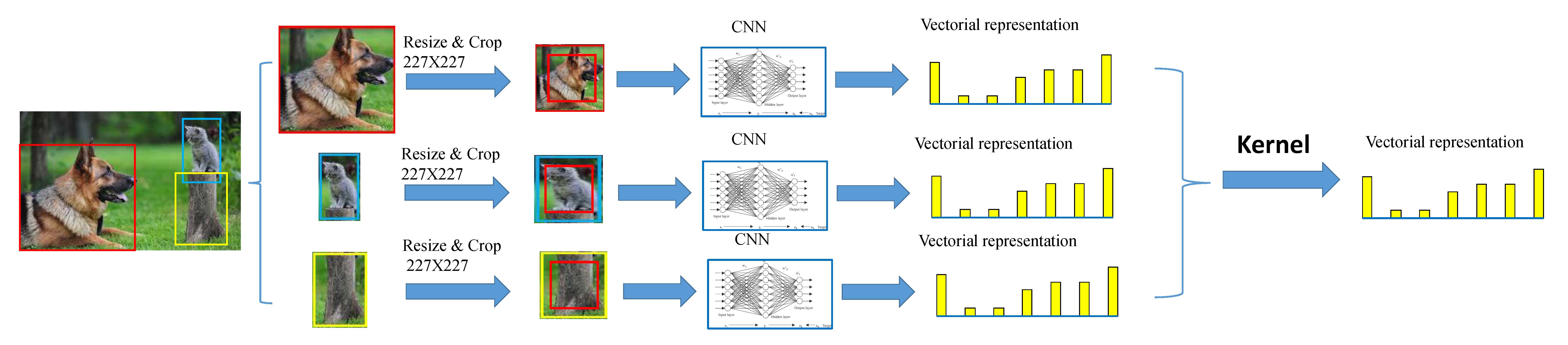}
  \label{cnn_s3}
 }
\end{center}
\caption{The illustration of our motivation to propose the kernelized convolutional neural network. (We refer the CNN details to the Caffe implementation. There should not be substantial differences from the original CNN model in \cite{krizhevsky2012imagenet}.) (a) A simple image with a single object localized at the center (roughly aligned) (b) A complex image with several objects (c) The proposed kernelized convolutional neural network algorithm}
\label{cnn_pipeline}
\end{figure*}

Kernel method is another alternative to transform a set of features into a vectorial representation, such as Fisher kernel \cite{FisherVector}, and  democratic kernel \cite{jegou:hal-00977321}. Fisher kernel models the joint probability distribution of the visual features detected in an image. The vectorial representation is constructed based on the derivatives in the parameter space. Besides the quantization results in the BoVW model, Fisher kernel also includes the residual information between the local visual features and their visual words \cite{jegou2012aggregating}. Fisher kernel is demonstrated to be more efficient than the BoVW model in image classification and image search applications \cite{FisherVector, jegou:hal-00977321, jegou2010aggregating, jegou2012negative, perronnin2010improving}. One non probabilistic version of Fisher kernel is carefully investigated in \cite{jegou2010aggregating, jegou2012aggregating}, which is named as vector of locally aggregated descriptors (VLAD).

Instead of designing the handcraft visual features, such as SIFT \cite{SIFT}, SURF \cite{bay2006surf}, and HOG \cite{HOG}, deep convolutional neural network (CNN) \cite{krizhevsky2012imagenet} learns a non-linear transformation model from large-scale well organized semantic dataset, namely ImageNet \cite{imagenet}. With the learned non-linear transformation model, each image can be transformed to a feature vector \cite{jia2014caffe}. With deep nets to learn from large-scale dataset, the CNN model can well discriminate diverse visual content, which is desired in many visual information processing systems. With breakthrough in many computer vision tasks, the CNN model has made a milestone in visual representation and become a new benchmark baseline \cite{razavian2014cnn}.

A lot of efforts have been made to understand the representation ability of convolutional neural network \cite{goodfellow2009measuring, zeiler2014visualizing, lenc2014understanding, cimpoi2015deep, long2014convnets, razavian2015persistent}. In \cite{goodfellow2009measuring}, Goodfellow et al. test the invariance of deep networks with a natural video dataset and find that the ``deep" structure can obtain more invariance than the ``shallow" ones. In \cite{zeiler2014visualizing}, Zeiler and Fergus try to understand why deep convolutional neural network works very well. They propose to visualize the patterns activated by the intermediate layers with a deconvolutional network. It is revealed that some complex patterns can be captured by top layers, which is very amazing. In \cite{lenc2014understanding}, Lenc et al. study the mathematical properties of equivariance, invariance, equivalence of image representations such as SIFT or CNN from the theoretical perspective. In \cite{cimpoi2015deep}, Cimpoi et al. conduct a range of experiments on material and texture attribute recognition and find that CNN can also obtain excellent result on this topic. In \cite{long2014convnets}, Long et al. study the learned correspondence at a fine level of CNN and reveal that good keypoint prediction can be obtained with the learned intermediate CNN features. More specifically, in \cite{razavian2015persistent}, Razavian et al. demonstrate that local spatial information of image is also conveyed by CNN and this local information can be used to perform facial landmark prediction, semantic segmentation, and object keypoints detection.

However, CNN is suitable to describe these images with a single object localized at the center, namely those roughly aligned images as shown in Fig. \ref{cnn_s1}. For a complex image with multiple objects, it is unsuitable to extract a single global CNN feature as shown in Fig. \ref{cnn_s2} because there may exits geometric transformations on these objects. As a more reasonable alternative, we can firstly align the content of the image and then construct the global vectorial representation. Hence, inspired by the invariant representation via pooling local features, in this paper we propose to represent image with local CNN to address the translation and re-sizing invariance issue and pool the transformed CNN feature to achieve a fix-sized rotation-invariant representation, which we call the kernelized convolutional neural network (KCNN) in the following. Specifically, we first detect some object-like patches from the given image. Then for each detected object-like patch, we extract CNN feature to describe the object in it. Finally to form a vectorial representation of the whole image, we aggregate these object-level CNN features with kernel function as shown in Fig. \ref{cnn_s3}.

We organize the rest of the paper as follows. In Section \ref{global_cnn_analysis}, we present some studies on the sensitivity of global CNN feature to three specific transformations. In Section \ref{KCNN}, we introduce our algorithm in detail. The experimental results are presented in Section \ref{Experimental_Results}. Finally we make conclusions in Section \ref{conclusion}.

\section{Sensitivity of Global CNN Feature} \label{global_cnn_analysis}
In this section, we study the sensitivity of global CNN feature to geometric transformations, \emph{i.e.}, translation, scaling, and rotation in detail. The study is made on the Holidays \cite{jegou2010aggregating} dataset which is a benchmark dataset for image search with 1491 high resolution images. We use the Caffe-based CNN implementation \cite{jia2014caffe} to extract our CNN feature. In the following, given an image $I$, we use $f(\cdot)$ to denote its extracted CNN feature in "fc7" layer and use $m(\cdot)$ to denote the cosine similarity between two CNN features.  All CNN feature are L2-normalized in default. To reveal the impact of geometric transformations to the global CNN feature independently, we design the following experiments to make sure that each image undergoes only one kind of geometric transformations.
\\

\begin{figure}[t]
\begin{center}
\subfigure[]{
\begin{minipage}[]{1.0 \linewidth}
\centering
  \includegraphics[width=1.0 \linewidth]{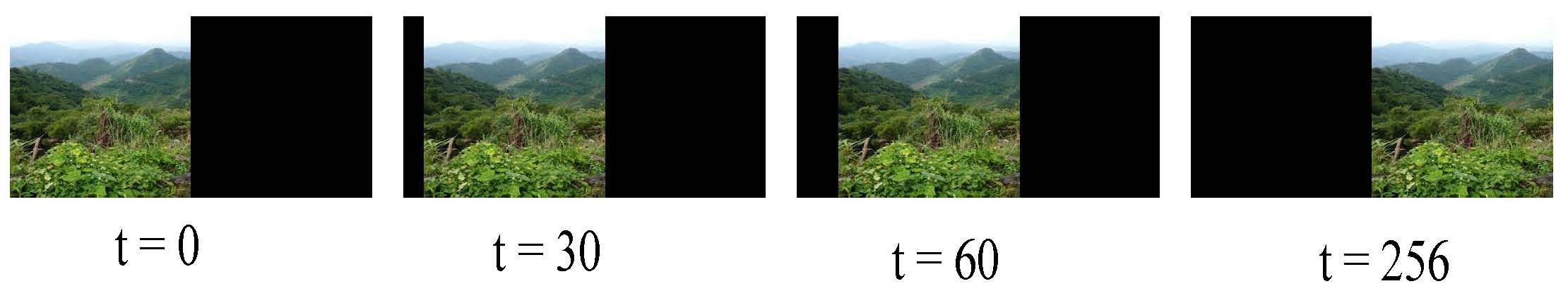}
\end{minipage}
\label{translation_design_a}
}
 \subfigure[]{
 \begin{minipage}[]{1.0 \linewidth}
 \centering
 \includegraphics[width=0.23 \linewidth, height=0.23 \linewidth]{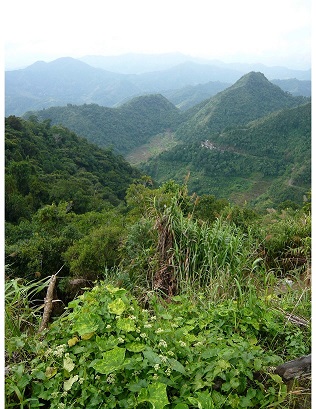}
 \includegraphics[width=0.23 \linewidth, height=0.23 \linewidth]{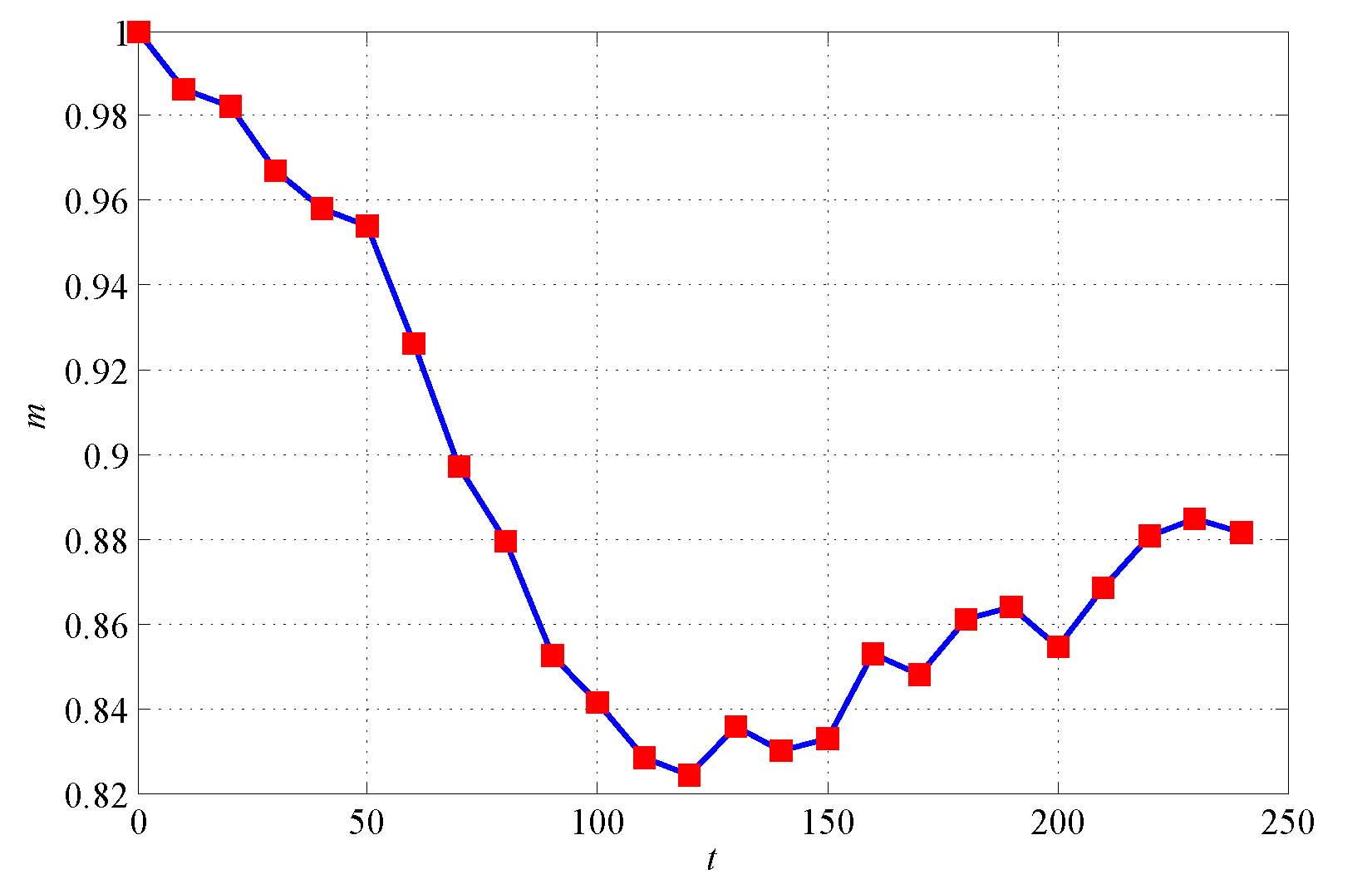}
 \includegraphics[width=0.23 \linewidth, height=0.23 \linewidth]{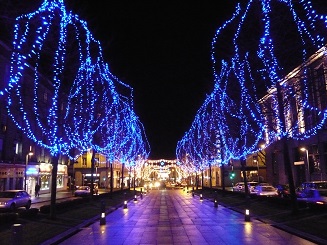}
 \includegraphics[width=0.23 \linewidth, height=0.23 \linewidth]{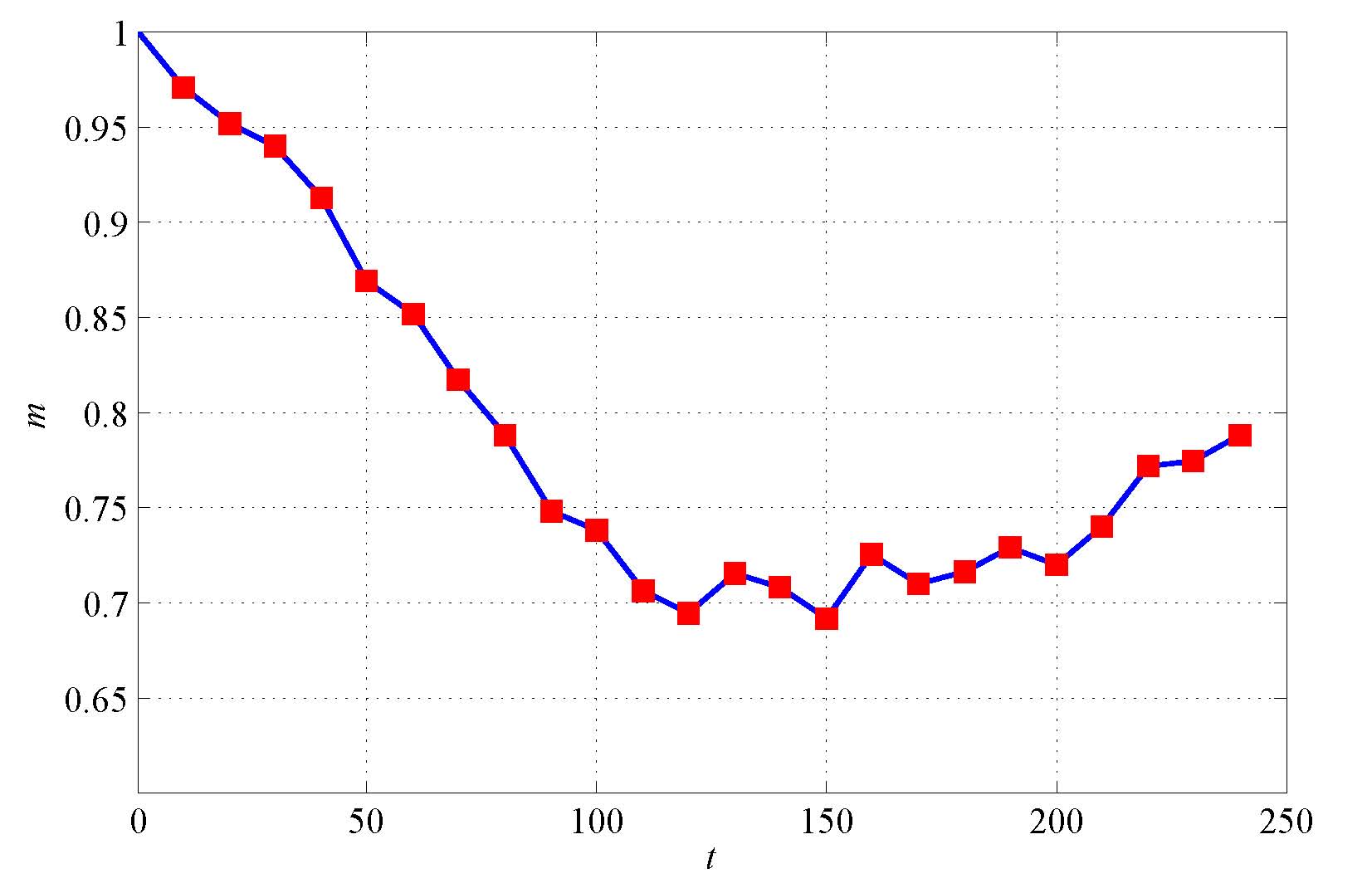}
 \end{minipage}
 \label{translation_design_b}
 }
 \subfigure[]{
 \begin{minipage}[]{0.8 \linewidth}
 \centering
 \includegraphics[width=0.5 \linewidth]{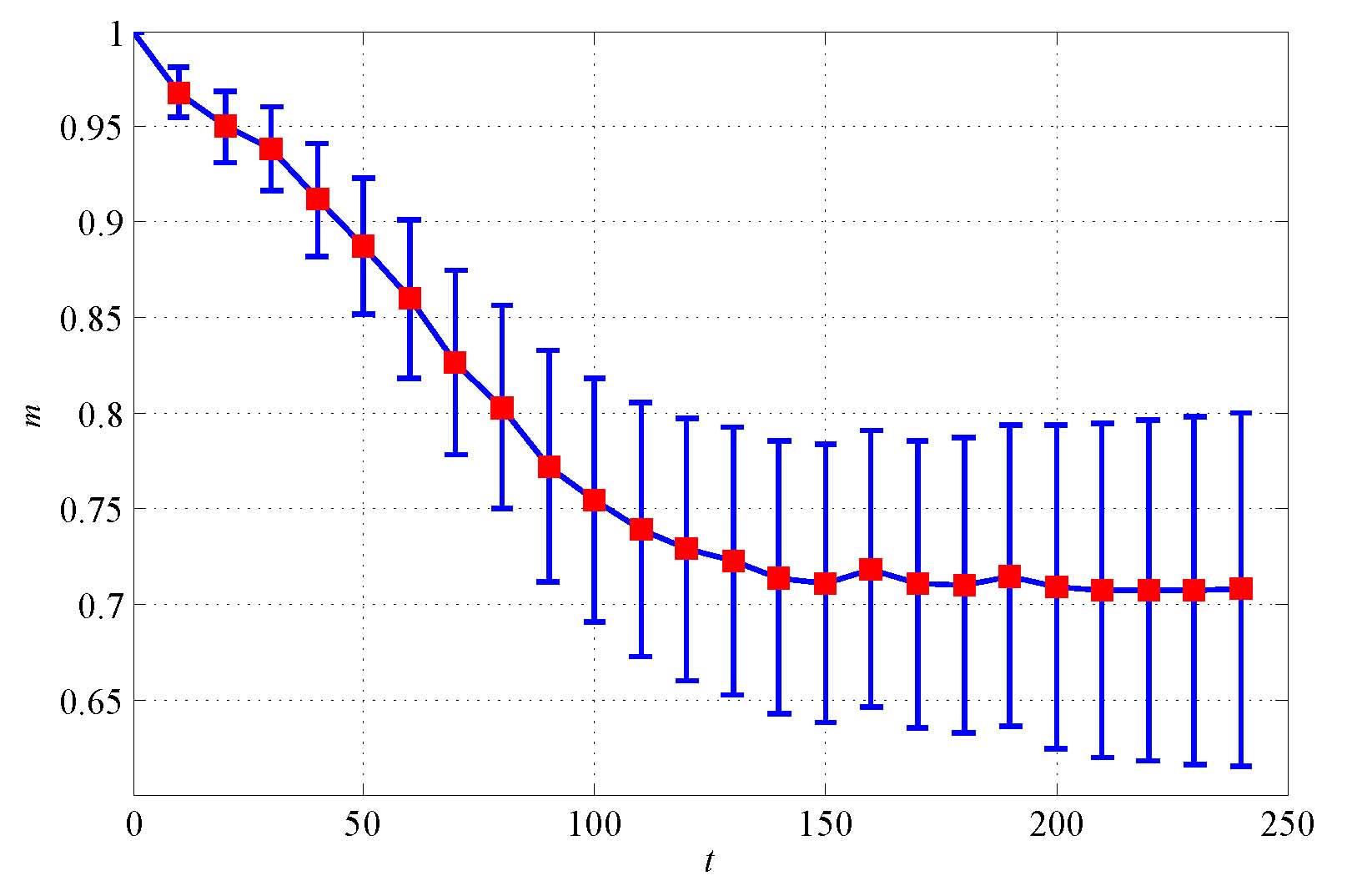}
 \end{minipage}
 \label{translation_design_c}
 }
\end{center}
\caption{The experiment to study the translation property of global CNN feature. (a) The illustration of image translation (b) Two examples of the similarities of the global CNN feature before and after the translation transformation (c) The mean and standard deviation of similarities of the global CNN features with respect to the translation transformation}
\label{translation_design}
\end{figure}

\noindent{\textbf{Translation.}} Generally, a translation can be made in vertical and horizontal directions. To simplify the study, we consider only the translation in the horizontal direction as shown in Fig. \ref{translation_design}. The extension to the general translation is straightforward. Given an image $I$ with size $M$ by $N$, we generate a larger image with size $M \times 2N$, as shown in Fig. \ref{translation_design_a} and pad the left half part with image $I$ by the border extrapolation method. Then we circularly translate $I$ by $t$ pixels to the left and construct its transformed version $I(t)$ and extract the global CNN feature $f(I(t))$. We measure the consistency score between global CNN features of $I(t=0)$ and $I(t)$ with their cosine similarity, as shown by the following equation.
\begin{equation} \label{translate_eqt}
m(I(t)) = <f(I(t=0)), f(I(t))>
\end{equation}
in which $<\cdot, \cdot>$ means the inner product operation.

In Fig. \ref{translation_design_b}, we illustrate two examples of the similarity between the global CNN features before and after the translation transformation. It can be seen that with the increase of horizontal translation, the similarity first declines and then grows after it reaches a valley. The decrease in similarity reflects the fact that the global CNN feature is sensitive to the translation transformation. On the other hand, the increase of the similarity after the valley point demonstrates the effect of the flipping operation which is make during the training stage of the CNN model. Similar phenomenon is also demonstrated by the statistical results shown in Fig. \ref{translation_design_c}. The difference in the trends of the similarity curves reflects the tolerance capability of global CNN feature to the translation transformation is also related to the content of image.
\\

\begin{figure}[t]
\begin{center}
\subfigure[]{
\begin{minipage}[]{1.0 \linewidth}
\centering
  \includegraphics[width=0.75 \linewidth]{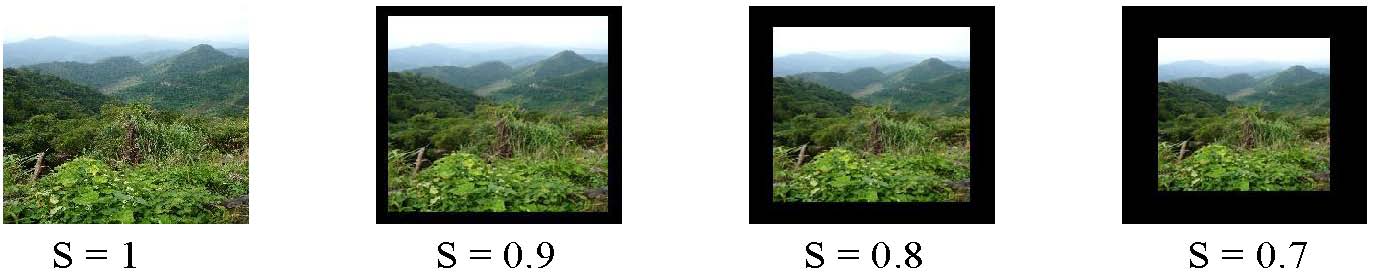}
\end{minipage}
\label{scaling_design_a}
}
 \subfigure[]{
 \begin{minipage}[]{1.0 \linewidth}
 \centering
 \includegraphics[width=0.2 \linewidth, height=0.2 \linewidth]{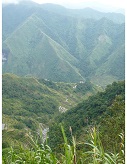}
 \includegraphics[width=0.2 \linewidth, height=0.2 \linewidth]{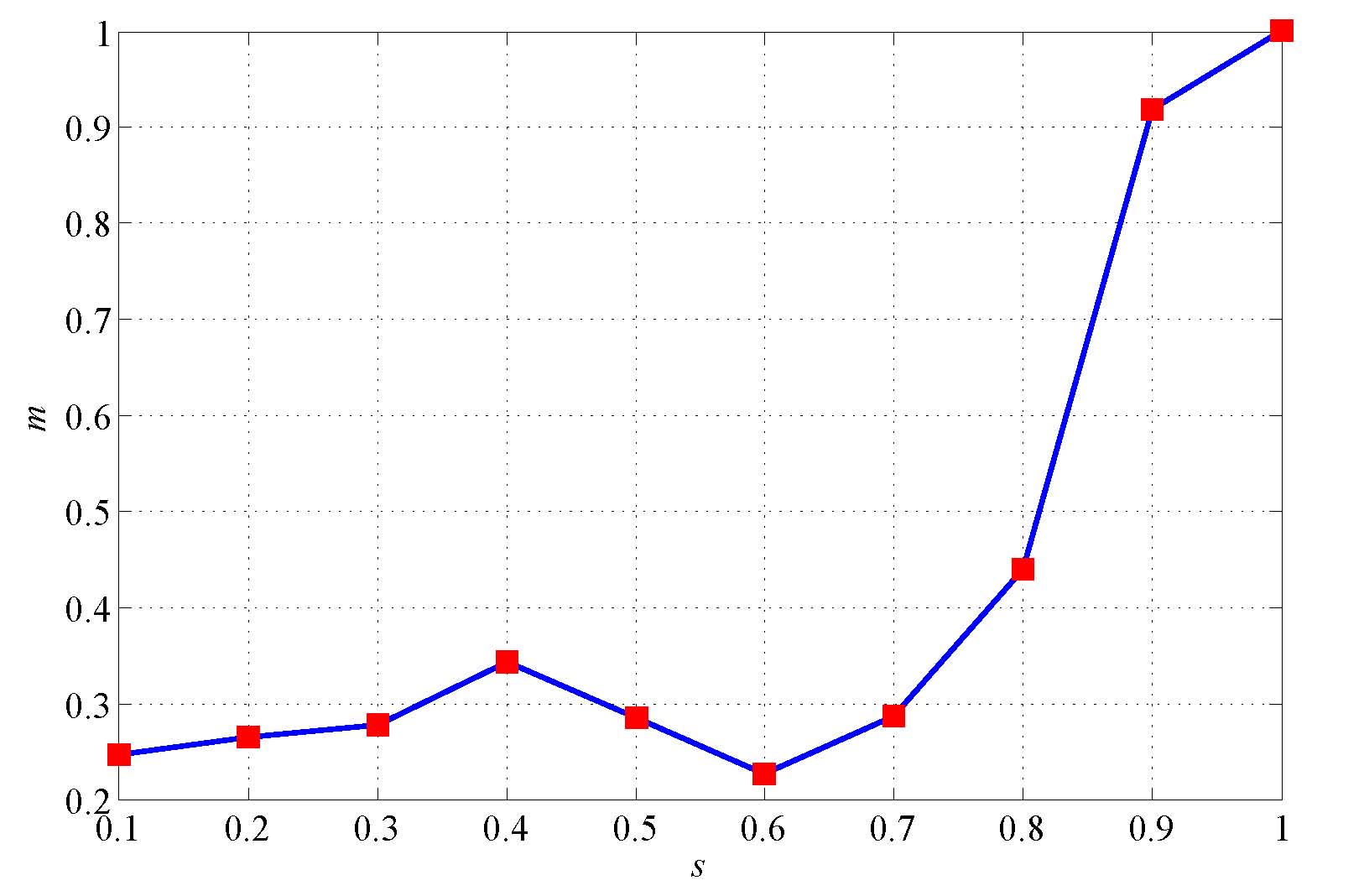}
 \includegraphics[width=0.2 \linewidth, height=0.2 \linewidth]{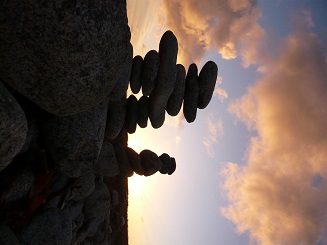}
 \includegraphics[width=0.2 \linewidth, height=0.2 \linewidth]{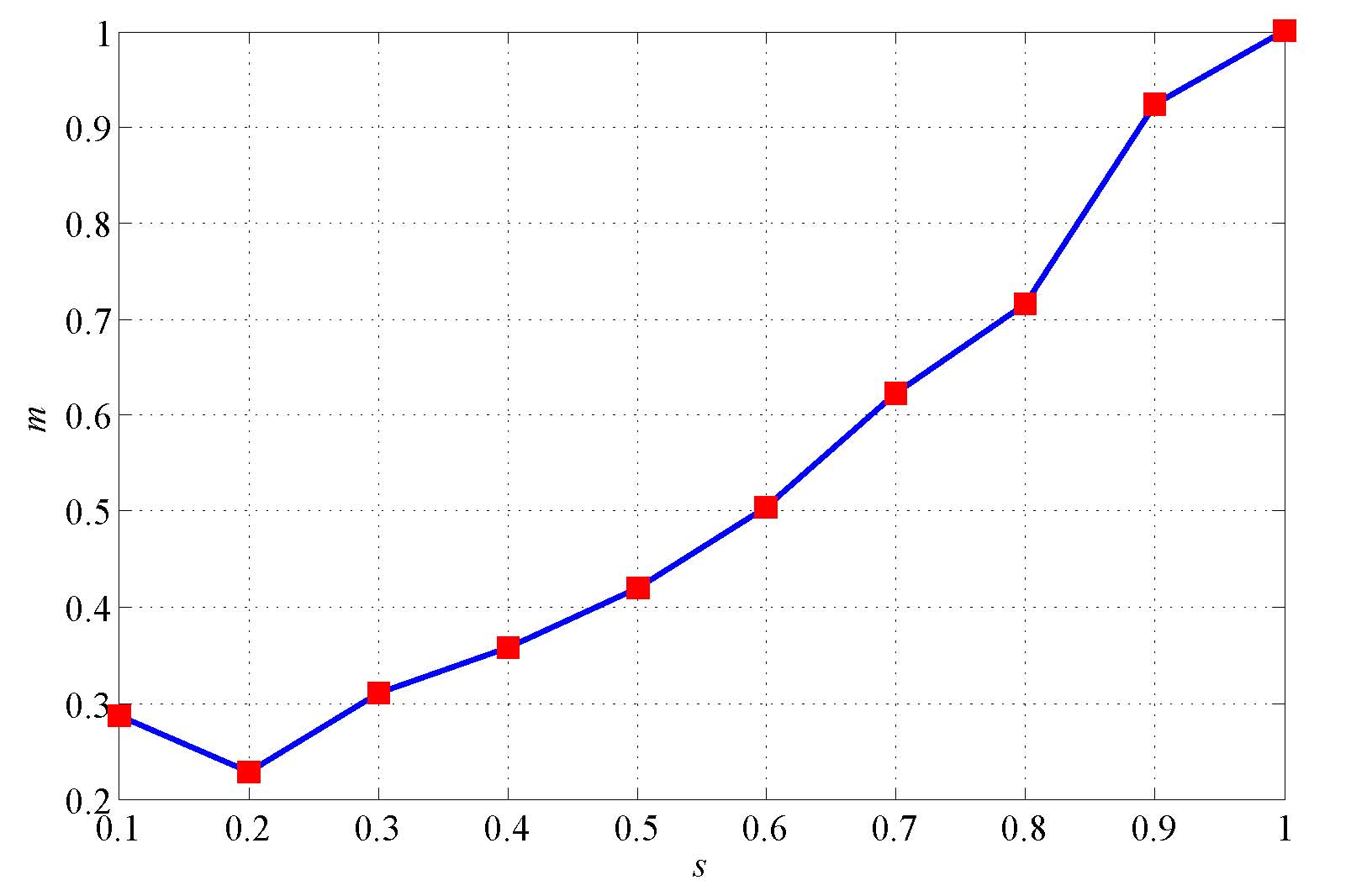}
 \end{minipage}
 \label{scaling_design_b}
 }
 \subfigure[]{
 \begin{minipage}[]{0.8 \linewidth}
 \centering
 \includegraphics[width=0.5 \linewidth]{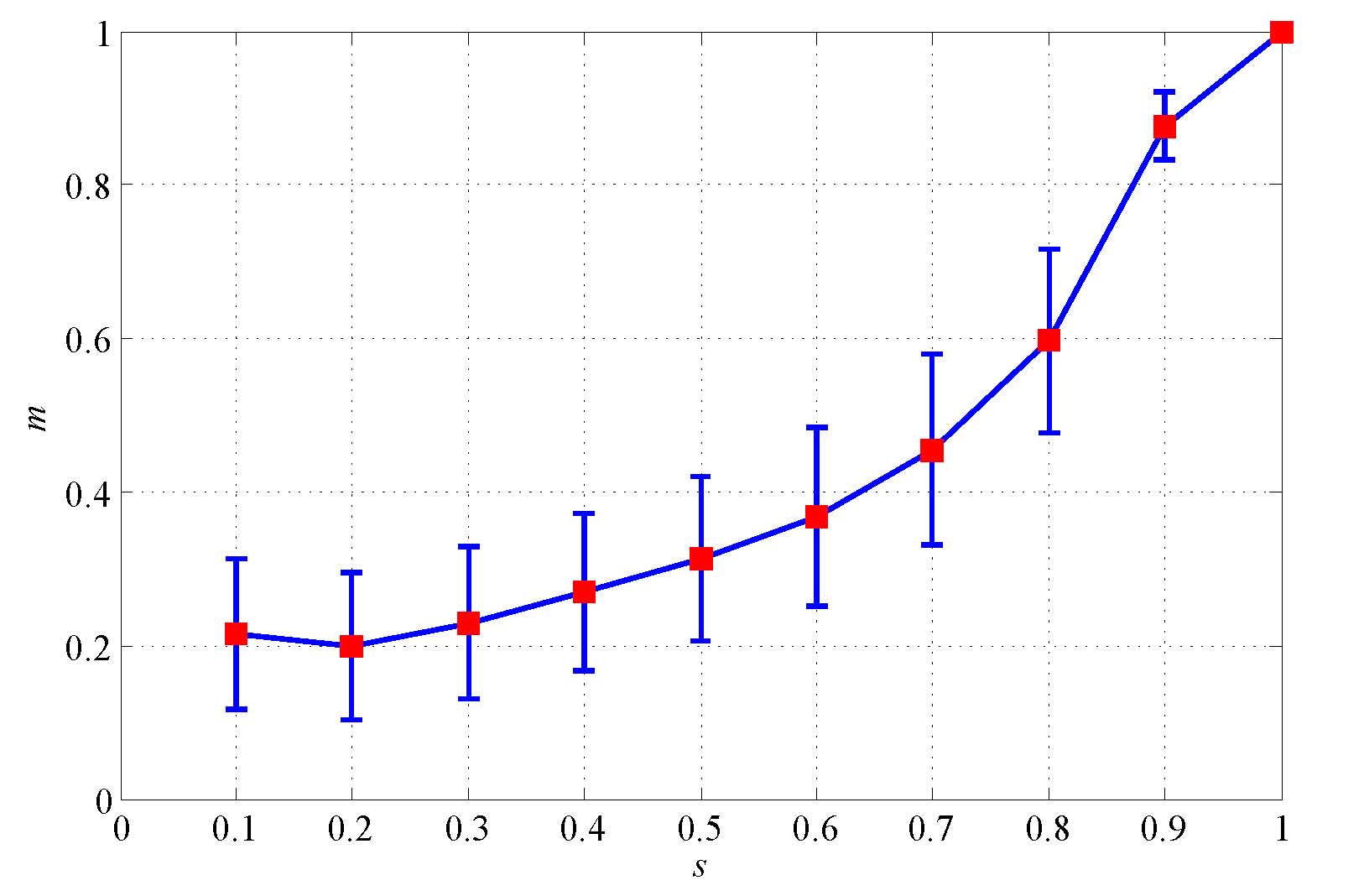}
 \end{minipage}
 \label{scaling_design_c}
 }
\end{center}
\caption{The experiment to study the scaling property of global CNN feature. (a) The illustration of image scaling (b) Two examples of the similarities of the global CNN feature before and after the scaling transformation (c) The mean and standard deviation of similarities of the global CNN features with respect to the scaling transformation}
\label{scaling_design}
\end{figure}

\noindent{\textbf{Scaling.}} In Fig. \ref{scaling_design}, we show our experiment to study the scaling property of the global CNN feature. The similarity to measure the image scaling transformation is defined as
\begin{equation}
m(I(s)) = <f(I(s=1)) , f(I(s))>,
\end{equation}
where $I(s)$ denotes the new image re-sized from the original image $I$ with the width and height being $s$ times of $I$. To keep the image $I(s)$ in the same size, we pad the region beyond the image boundary by the border extrapolation method. Another choice is to crop the sub-images different size at the same location.
However, there should not be substantial difference between these two methods to construct $I(s)$. Then we extract the global CNN feature $f(I(s))$.

In Fig. \ref{scaling_design_b}, we illustrate two examples of the similarity of the global CNN features before and after the scaling transformation. It can be seen that the similarity score decreases as the image is scaled with different ratios, which means the global CNN feature is not invariance to the scaling transformation. Similar phenomenon is also demonstrated by the statistical results shown in Fig. \ref{scaling_design_c}.
\\

\begin{figure}[t]
\begin{center}
\subfigure[]{
\begin{minipage}[]{1.0 \linewidth}
\centering
  \includegraphics[width=1.0 \linewidth]{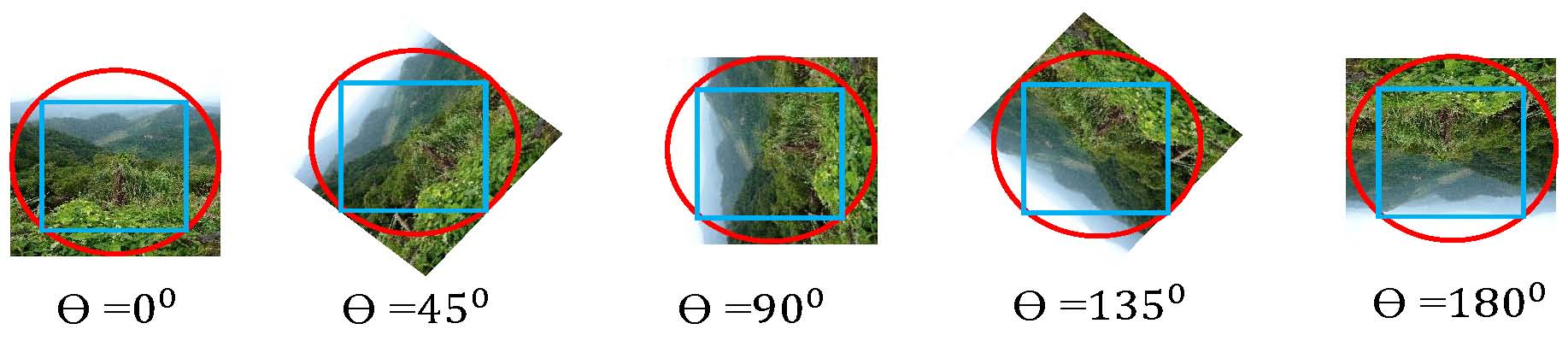}
\end{minipage}
\label{rotation_design_a}
}
 \subfigure[]{
 \begin{minipage}[]{1.0 \linewidth}
 \centering
 \includegraphics[width=0.2 \linewidth, height=0.2 \linewidth]{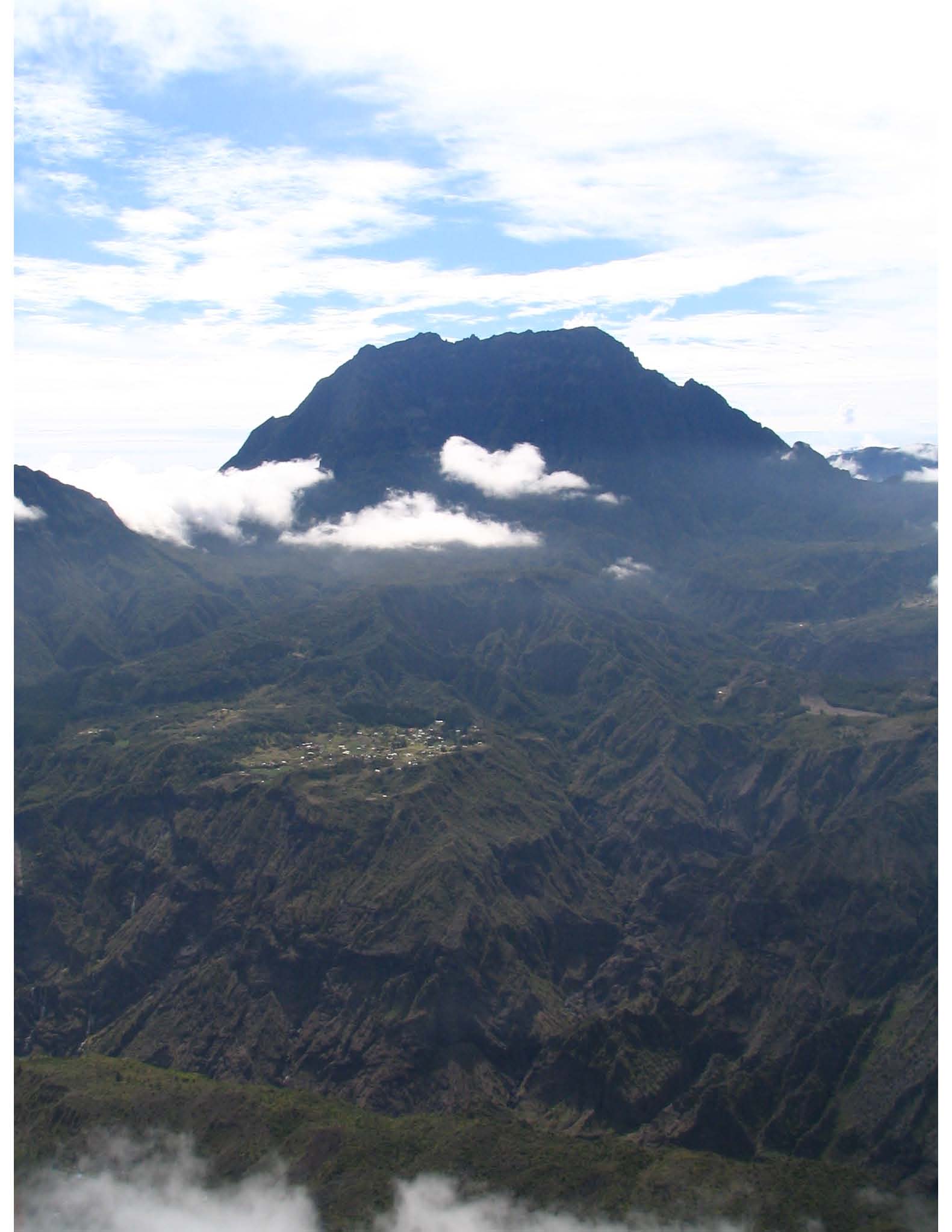}
 \includegraphics[width=0.2 \linewidth, height=0.2 \linewidth]{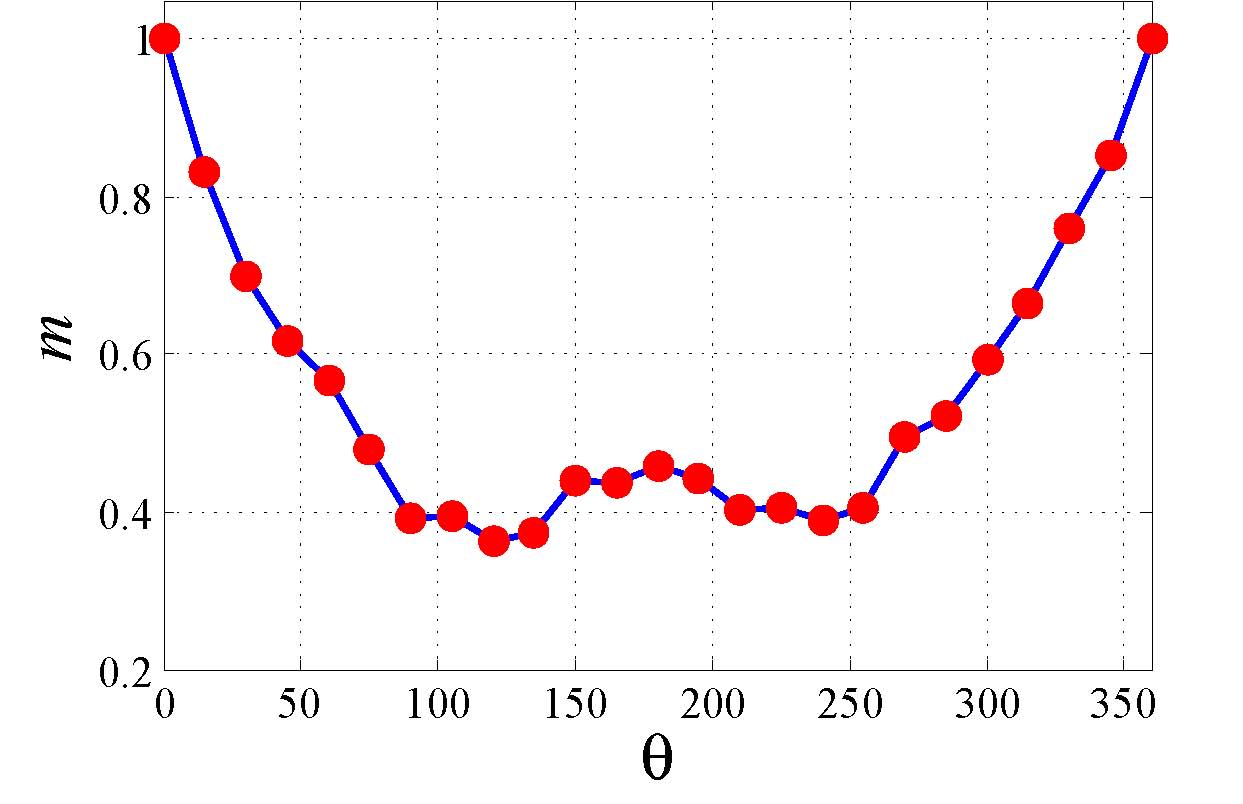}
 \includegraphics[width=0.2 \linewidth, height=0.2 \linewidth]{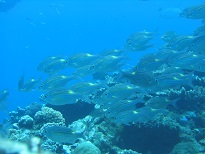}
 \includegraphics[width=0.2 \linewidth, height=0.2 \linewidth]{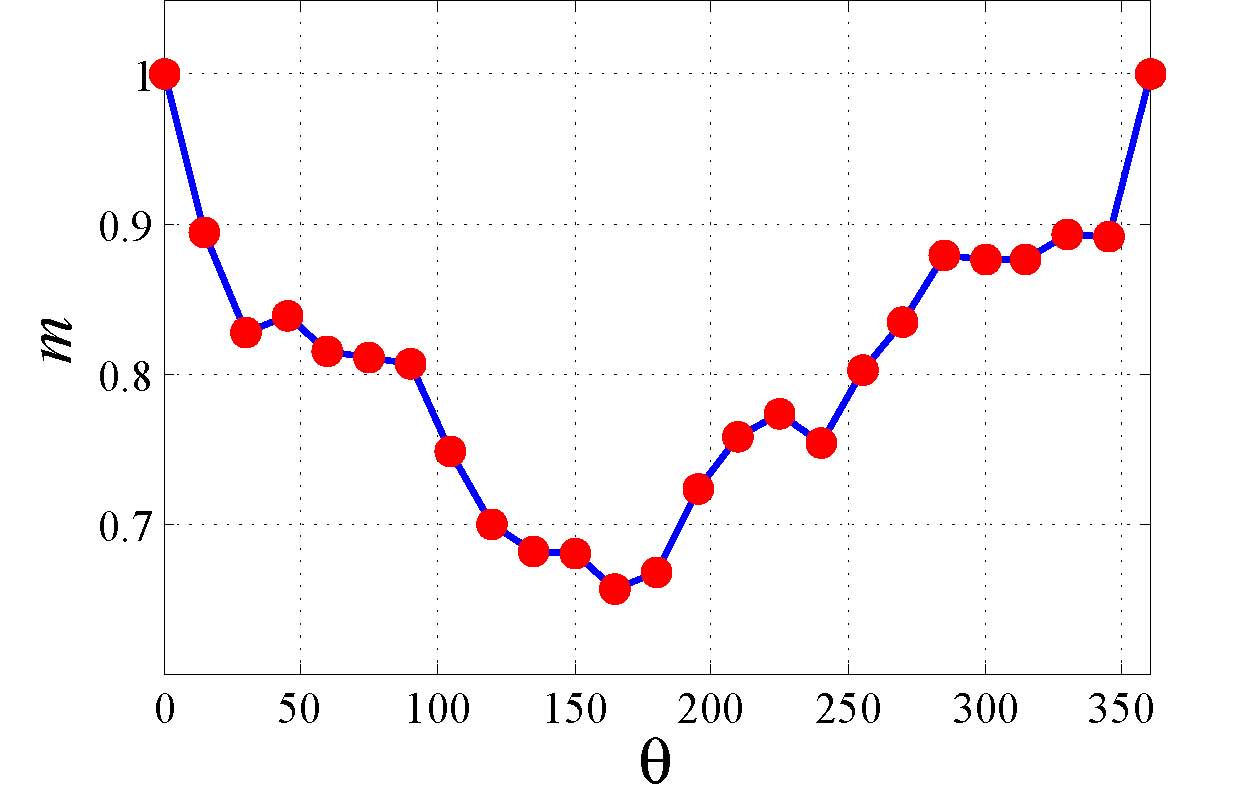}
 \end{minipage}
 \label{rotation_design_b}
 }
 \subfigure[]{
 \begin{minipage}[]{0.8 \linewidth}
 \centering
 \includegraphics[width=0.5 \linewidth]{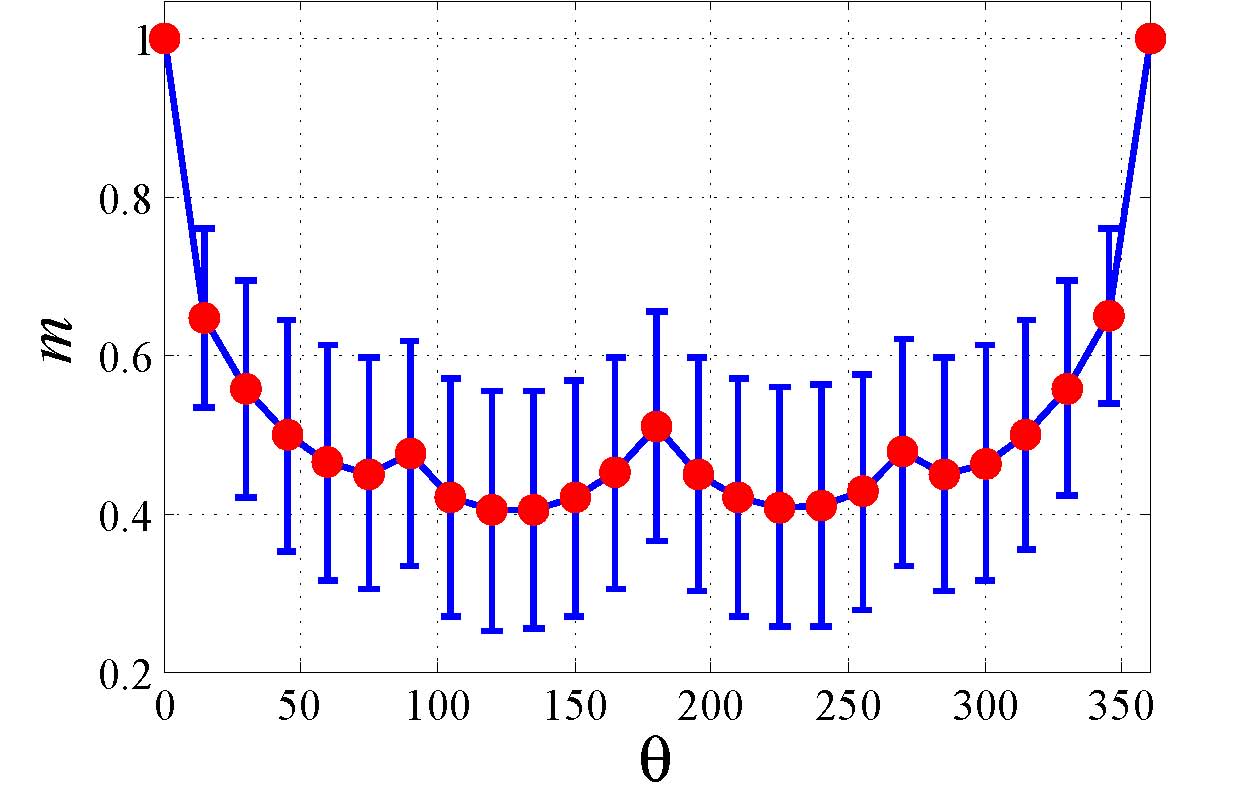}
 \end{minipage}
 \label{rotation_design_c}
 }
\end{center}
\caption{The experiment to study the rotation property of global CNN feature. (a) The illustration of image rotation (b) Two examples of the similarities of the global CNN feature before and after the rotation transformation (c) The mean and standard deviation of similarities of the global CNN features with respect to the rotation transformation}
\label{rotation_design}
\end{figure}

\noindent{\textbf{Rotation.}} In Fig. \ref{rotation_design}, we show our experiment to study the rotation property of the global CNN feature.  We measure the consistency score of CNN feature to rotation transformation as
\begin{equation}
m(I(\theta)) = <f(I(\theta=0\text{\degree)}) , f(I(\theta))>,
\end{equation}
where $I(\theta)$ denotes the new image after the image $I$ is rotated by $\theta$ degree, as shown in Fig. \ref{rotation_design_a}. Please note that the image size will change after rotation as shown by comparing the figure of $\theta = 0$\degree and the figure of $\theta = 45$\degree in Fig. \ref{rotation_design_a}. To study the property of global CNN feature when only rotation transformation exists, we extract the CNN feature on the sub-image located at the center of $I(s)$ as illustrated in the blue square of the red inscribed circle in Fig. \ref{rotation_design_a}.

In Fig. \ref{rotation_design_b}, we illustrate two examples of the similarity of the global CNN features before and after the rotation transformation. It can be seen that the similarity varies as the image is rotated with different degrees and the similarity curves of these two examples have different trends. Similar phenomenon is also demonstrated by the statistical results shown in Fig. \ref{rotation_design_c}, which demonstrated that the global CNN feature is sensitive to the rotation transformation. That the similarity curves have different trends means the tolerance ability to the rotation transformation of global CNN feature is also related to the content of image.
\\

\noindent{\textbf{Discussion.}} From the experiments above, it can be observed that the similarity $m$ of the global CNN features before and after transformation is sensitive to translation, rotation, and scaling. This comes from the architecture of the CNN model in which the neurons are highly related to the spatial positions of the image pixels in local perception field. When the image is transformed, the spatial positions of those pixels are changed, which results in the inconsistent CNN feature and limits the robustness of CNN feature to these geometric transformations such as translation, scaling, and rotation. To address this problem we propose to firstly align the image content in the patch level before extracting the CNN feature. Such a strategy makes the feature robust to translation and scaling change. Moreover, to enhance the robustness to rotation changes, each image patch is rotated circularly by 8 times. Then to build a vectorial image level representation, we aggregate the extracted patch-level CNN features with kernel functions.

\section{Kernelized Convolutional Neural Network} \label{KCNN}

\begin{figure}[t]
\begin{center}
\subfigure[]{
\begin{minipage}[]{0.8 \linewidth}
\centering
 \includegraphics[width=0.45 \linewidth]{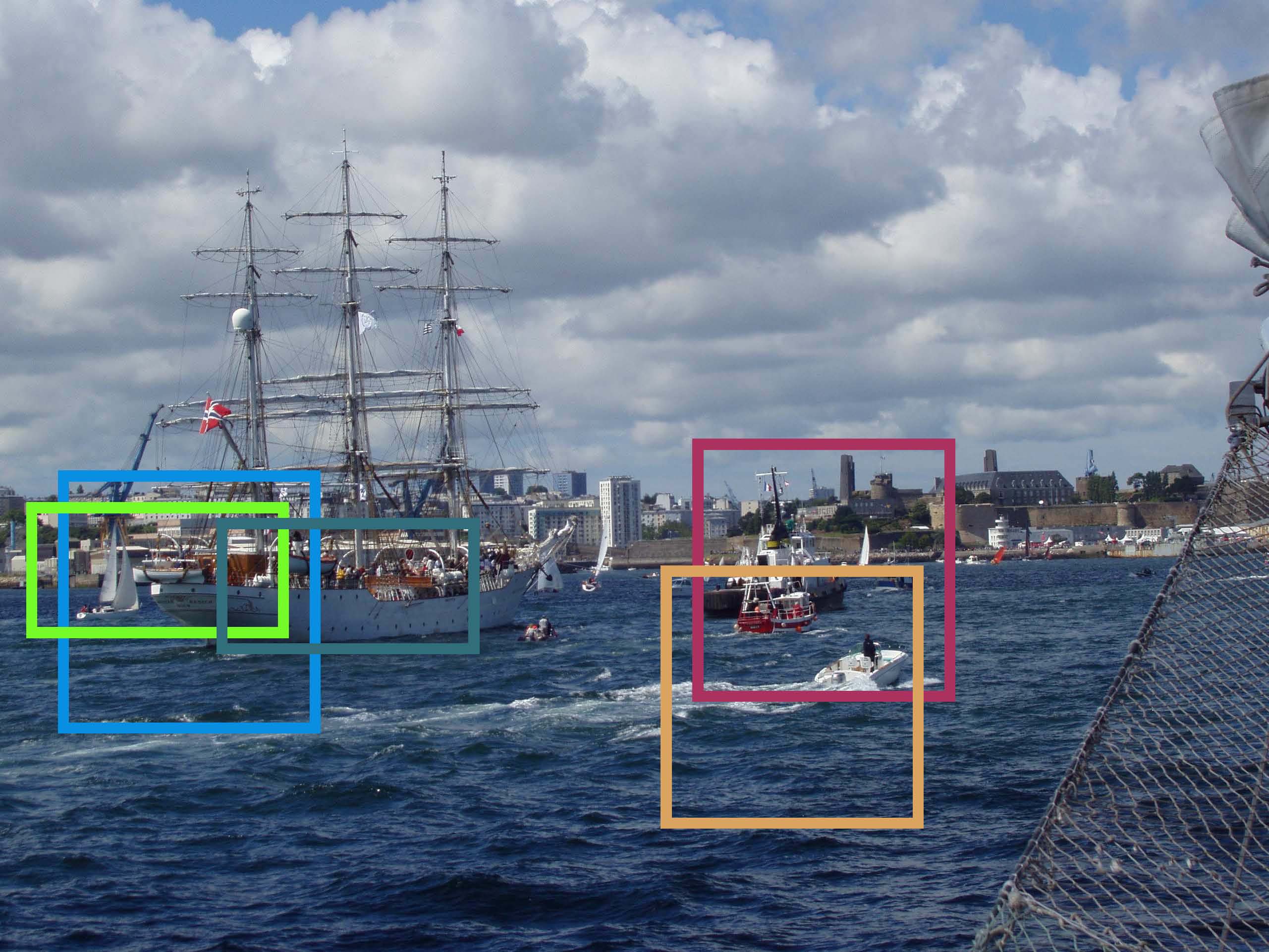}
 \includegraphics[width=0.45 \linewidth]{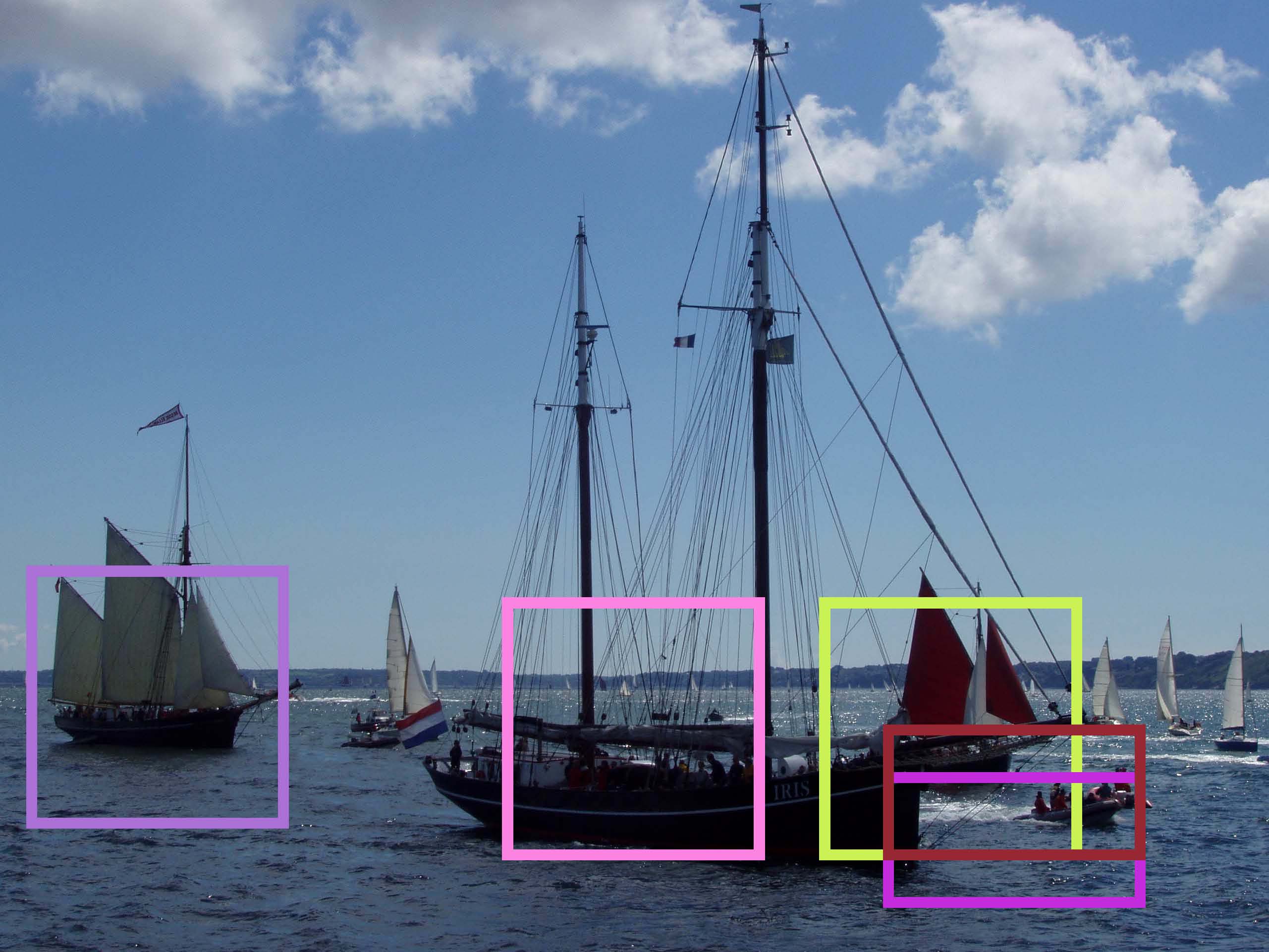}
\end{minipage}}
\end{center}
\caption{ Two example images with detected object-like patches. Only several top ranked patches are shown.}
\label{detect_patches}
\end{figure}

In this section, we introduce our algorithm to construct the vectorial representations on the roughly content-aligned images with the kernel method and the deep convolutional neural network in detail.

Given two sets of image patches $\mathcal{X}$ and $\mathcal{Y}$ with $\text{card($\mathcal{X}$)} = n \text{ and } \text{card($\mathcal{Y}$)} = m$. Let's consider using match kernel $\mathcal{K}(\cdot, \cdot)$ \cite{nips1988fisher, bo_nips09, jegou:hal-00977321} to measure the similarity between $\mathcal{X}$ and $\mathcal{Y}$, hence we have
\begin{equation} \label{e_kernel}
\mathcal{K}(\mathcal{X}, \mathcal{Y}) = \sum_{x \in \mathcal{X}}{\sum_{y \in \mathcal{Y}}{k(x, y)}},
\end{equation}
where $k(\cdot, \cdot)$ measures the similarity between two feature descriptors and $x$ stands for an image patch and $y$ has the similar meaning.

To construct a vectorial image representation for each image, we consider these separable kernel functions. Namely the similarity between two feature descriptors $k(x, y)$ can be computed by the inner product operation, as shown by the following equation
\begin{equation} \label{kcnn_equation}
\begin{aligned}
\mathcal{K}(\mathcal{X}, \mathcal{Y}) =& \sum_{x \in \mathcal{X}}{\sum_{y \in \mathcal{Y}}{k(x, y)}} \\
=& \sum_{x \in \mathcal{X}}{\sum_{y \in \mathcal{Y}}{<\phi(x), \phi(y)>}} \\
=& <(\sum_{x \in \mathcal{X}}{\phi(x)}), (\sum_{y \in \mathcal{Y}}{\phi(y)})> \\
=& <\Psi(\mathcal{X}), \Psi(\mathcal{Y})>,
\end{aligned}
\end{equation}
where $\phi(\cdot)$ means a kind of linear or nonlinear transformation and $\Psi(\mathcal{X})$ is the final image-level vectorial representation we need.

In Eq. \ref{kcnn_equation}, the key issue is how to define the function $\phi(\cdot)$. Firstly, as the size of $x$ is not fixed, we need a function to transform $x$ into a fixed dimensional vectorial representation, which can be denoted by $\gamma(\cdot)$. Secondly, to aggregate these patch-level vectorial representations $\gamma(x)$ into the final image-level vectorial representation $\Psi(\mathcal{X})$, we need a function to map $\gamma(x)$ into another space. This step can be denoted by $\beta(\cdot)$. Such that we have the form
\begin{equation} \label{KCNN_e}
 \phi(x) = \beta(\gamma(x)).
\end{equation}
In the following, we will discuss how to design the function $\gamma(\cdot)$ and $\beta(\cdot)$.
\\

\noindent{$\boldsymbol{\gamma(\cdot)}$:} In computer vision, it is a fundamental problem to describe an image patch of various sizes into a fixed-length feature vector. There are many classic works on it \cite{SIFT, HOG, bay2006surf}. For example, in SIFT \cite{SIFT} algorithm, the spatially constrained gradient histogram is used to represent the image patch. With the development of the technology, some researchers turn to the large-scale machine learning techniques. The recently research works revealed that the deep convolutional neural network (CNN) is very powerful for many computer vision tasks \cite{razavian2014cnn}. The CNN model is learned from a million-scale database, ImageNet. With the advantage of the non-linearity and large number of parameters, CNN can easily handle the immense variants of vision tasks. In this paper, we adopt the CNN model \cite{jia2014caffe} to transform the image patch into its vectorial representation. In \cite{jia2014caffe}, a pre-trained CNN model and well organized code are provided to be publicly available for academic uses. We adopt the CNN model to obtain the vectorial representation of each image patch.
\\

\noindent{$\boldsymbol{\beta(\cdot)}$:} After the image patches are transformed into vectorial representations, we adopt the separable kernel methods to aggregate them together to represent the image. There are also a lot of works devoted to kernel methods \cite{nips1988fisher, bo_nips09, jegou:hal-00977321, perronnin2010improving, bo_cvpr11, FisherVector}. One classic separable kernel is the Fisher kernel which models the joint probability distribution of a set of features \cite{perronnin2010improving, FisherVector, Perronnin2007cvpr, jegou2010aggregating}. Perronnin \emph{et.al.} \cite{Perronnin2007cvpr, perronnin2010improving} applied Fisher kernel to image classification and image retrieval applications. They model the features' joint probability distribution with a Gaussian mixture (GMM) model. In Fisher kernel, the mapping function $\beta(\cdot)$ corresponds to the gradient function of the features' joint probability distribution with respect to the parameters of this distribution, scaled by the inverse square root of the Fisher information matrix. It gives the direction in parameter space into which the learned distribution should be modified to better fit the observed data. In comparison with the BoVW model, the Fisher kernel model can obtain higher accuracy. Hence given a set of features, we adopt the Fisher kernel to construct their vectorial representation.
\\ 

\noindent{$\boldsymbol{x}$.} To analyze the visual content in a given image, researchers usually extract some interesting patches from it. The word ``interesting" means some clearly defined rules, which can make the detected patches have the desired properties. For example, in SIFT algorithm \cite{SIFT}, the image patches are detected with different of Gaussian (DoG) method to obtain the scale invariant property. Then to obtain the rotation invariant property, the detected patches are aligned with the dominant orientation of its gradients. In this paper, we use the object detector \cite{BingObj2014, objectNess, objectAlexe2010, EndresECCV2010} to extract some object-like patches from the image. After some object-like patches are detected, these patches are spatially aligned, which can provide the property of invariance to translation and scaling transformations. In a most recently published work named BING object detector \cite{BingObj2014}, Cheng et al. proposed a very efficient algorithm to detect object-like image patches with a quite higher detection rate, which can process 300 frames per second on a single CPU. BING object detection algorithm \cite{BingObj2014} output a real value for each patch to indicate how the detected image patch is like to be an object. With this real value, we can control the number of image patches we want. Considering the excellent speed of BING algorithm, we adopt it \cite{BingObj2014} to extract our image patches. To achieve the rotation invariant property of the extracted image patches, we rotate each image patch, $x$, by 8 discrete degrees which consist of 0\degree, 45\degree, 90\degree, 135\degree, 180\degree, 225\degree, 270\degree, 315\degree. Intuitively, a dominant angle for each object patch can be estimated in the similar way as SIFT. However, our study reveals that such a strategy yields low performance, due to the unreliability of the dominant angle estimation in object-patch level. Some examples of detected object-like patches with BING algorithm are shown in Fig. \ref{detect_patches}.

\noindent{\textbf{Time cost}.} Besides the time cost to extract object-like patches with BING detector and the aggregation cost with Fisher kernel, the time to extract KCNN will be $8 \times N$ times of regular CNN, where $N$ means the number of detected objects. But, this can be accelerated with GPU clusters. Since our paper focus on addressing the sensitivity of regular CNN, in our implementation we use the CPU mode of Caffe library. To fairly show the effectiveness of KCNN, we use the linear search method to search the database with the inner product operation to compute the similarity of two images. Therefore the complexity will depend on the dimension of image vectorial representation.

\section{Experimental Results} \label{Experimental_Results}
In this section, we evaluate our algorithm on the image retrieval application. We adopt three public available benchmark datasets, \emph{i.e}, Holidays \cite{jegou2010aggregating} and UKBench \cite{vocabuary:tree} and Oxford Building \cite{all_vlad}, to demonstrate the impact of the parameters in our algorithm. We also compare our algorithm with some other methods for image retrieval application.

Holidays dataset \cite{jegou2010aggregating} contains 1491 high-resolution images of different scenes and objects with 500 queries. To evaluate the performance we use the average precision measure computed as the area under the precision-recall curve for a query. We compute the mean of the average precision for all queries to obtain a mean Average Precision (mAP) score, which is used to evaluate the overall performance \cite{AKM}.

UKBench dataset \cite{vocabuary:tree} contains 2550 objects or scenes, each with four images taken under different views or imaging conditions, resulting in 10200 images in total. In terms of accuracy measurement, the top-4 accuracy \cite{vocabuary:tree} is used as evaluation metric. For top-4 accuracy, for each query, the retrieval accuracy is measured by counting the number of correct images in top-4 returned results. Then the retrieval performance is averaged over all test queries.

Oxford Building dataset \cite{all_vlad, AKM} consists of 5062 images of buildings and 55 query images corresponding to 11 distinct buildings in Oxford. Images are annotated as either relevant, not relevant, or \emph{junk} indicating that it is unclear whether a user would consider the image as relevant or not. Following the recommended protocol, the \emph{junk} images are removed from the ranking results. The retrieval performance is also measured by the mean Average Precision (mAP) computed over the 55 queries.

Our experiments are implemented on a server with 32GB memory and 2.4GHz CPU of Intel Xeon.

\subsection{Impact of Parameters}

\begin{figure}[t]
\begin{center}
 \subfigure[$D$=32]{
 \includegraphics[width=0.3 \linewidth]{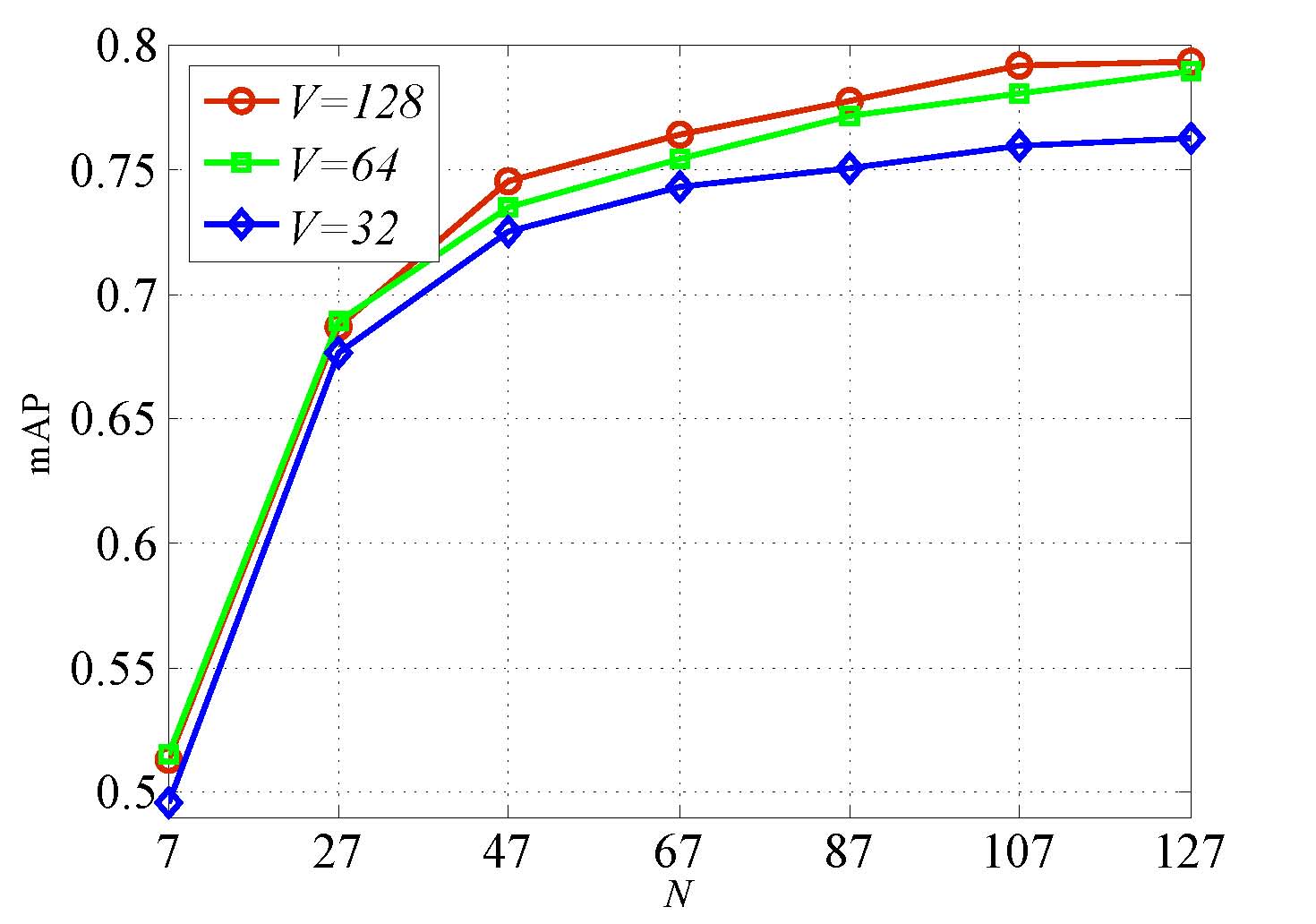}}
 \subfigure[$D$=64]{
 \includegraphics[width=0.3 \linewidth]{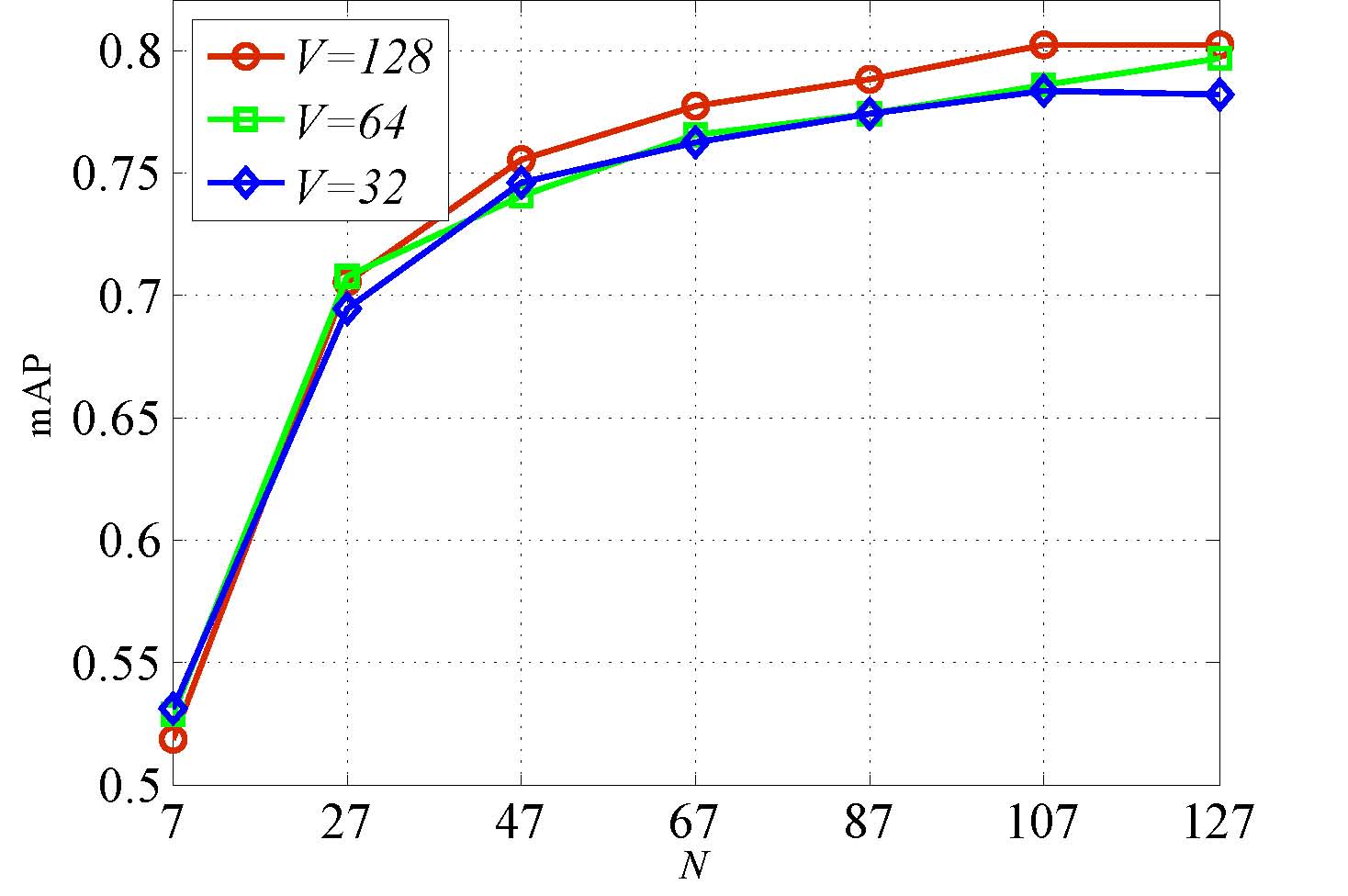}}
 \subfigure[$D$=128]{
 \includegraphics[width=0.3 \linewidth]{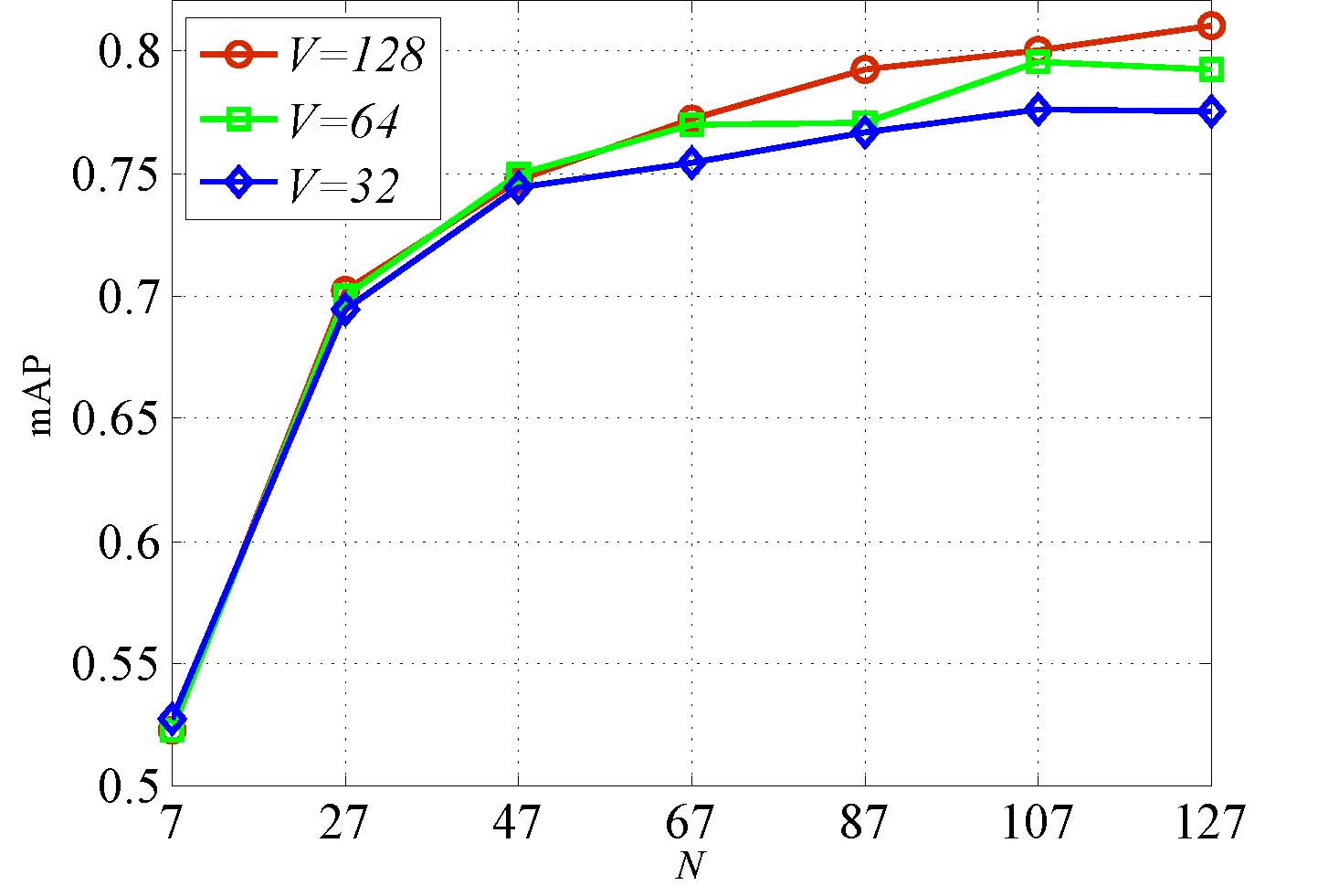}}
 \subfigure[$D$=32]{
 \includegraphics[width=0.3 \linewidth]{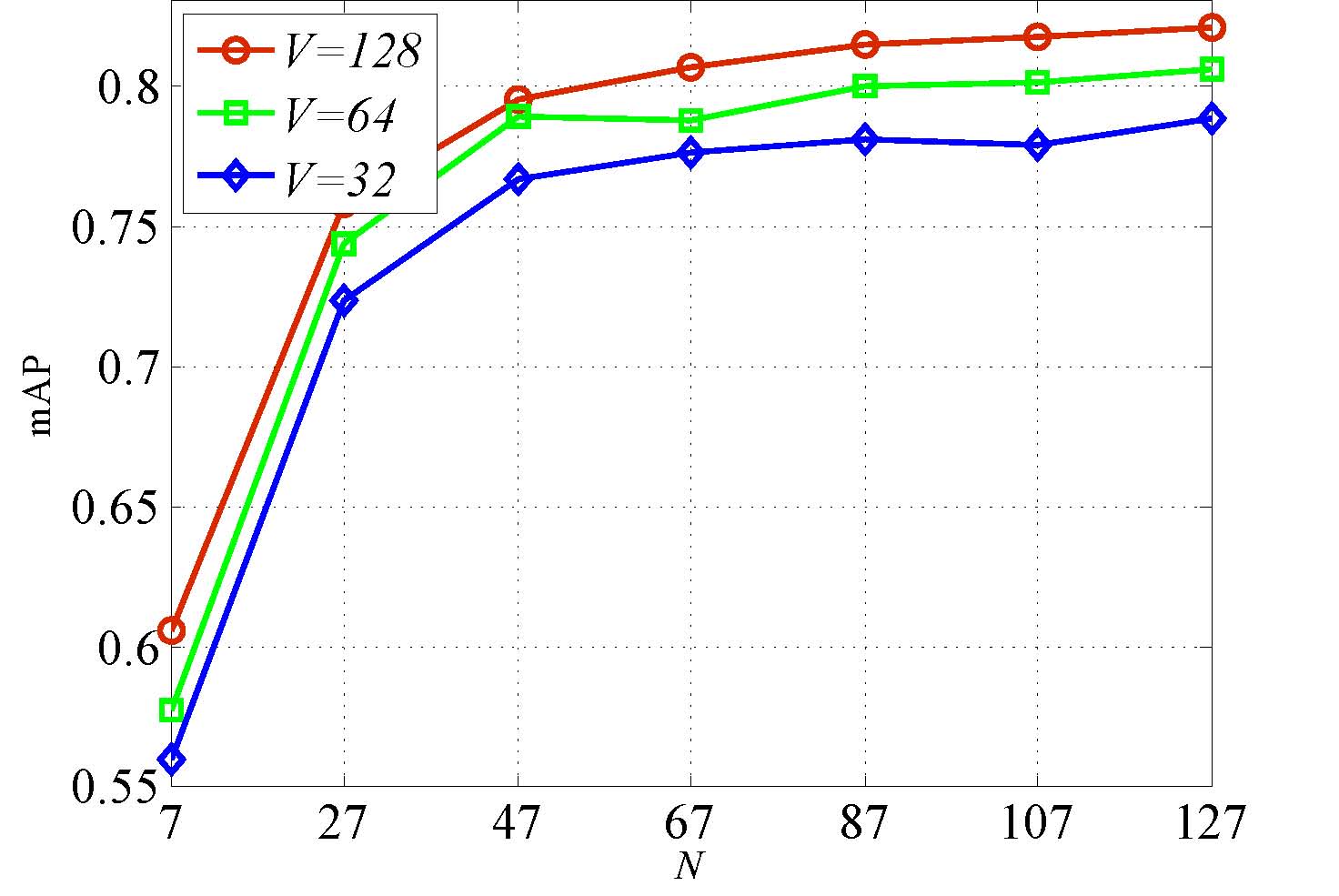}}
 \subfigure[$D$=64]{
 \includegraphics[width=0.3 \linewidth]{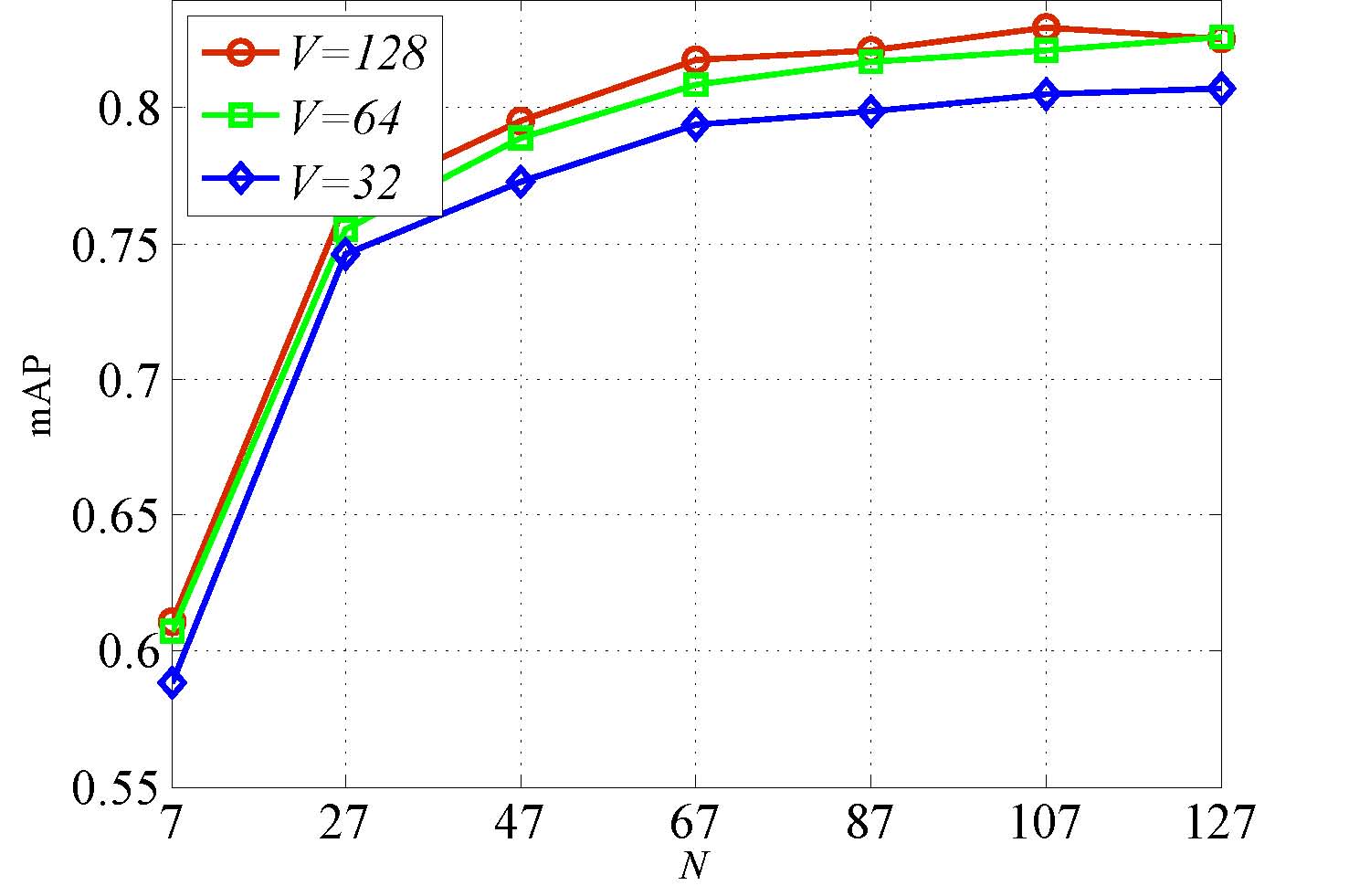}}
 \subfigure[$D$=128]{
 \includegraphics[width=0.3 \linewidth]{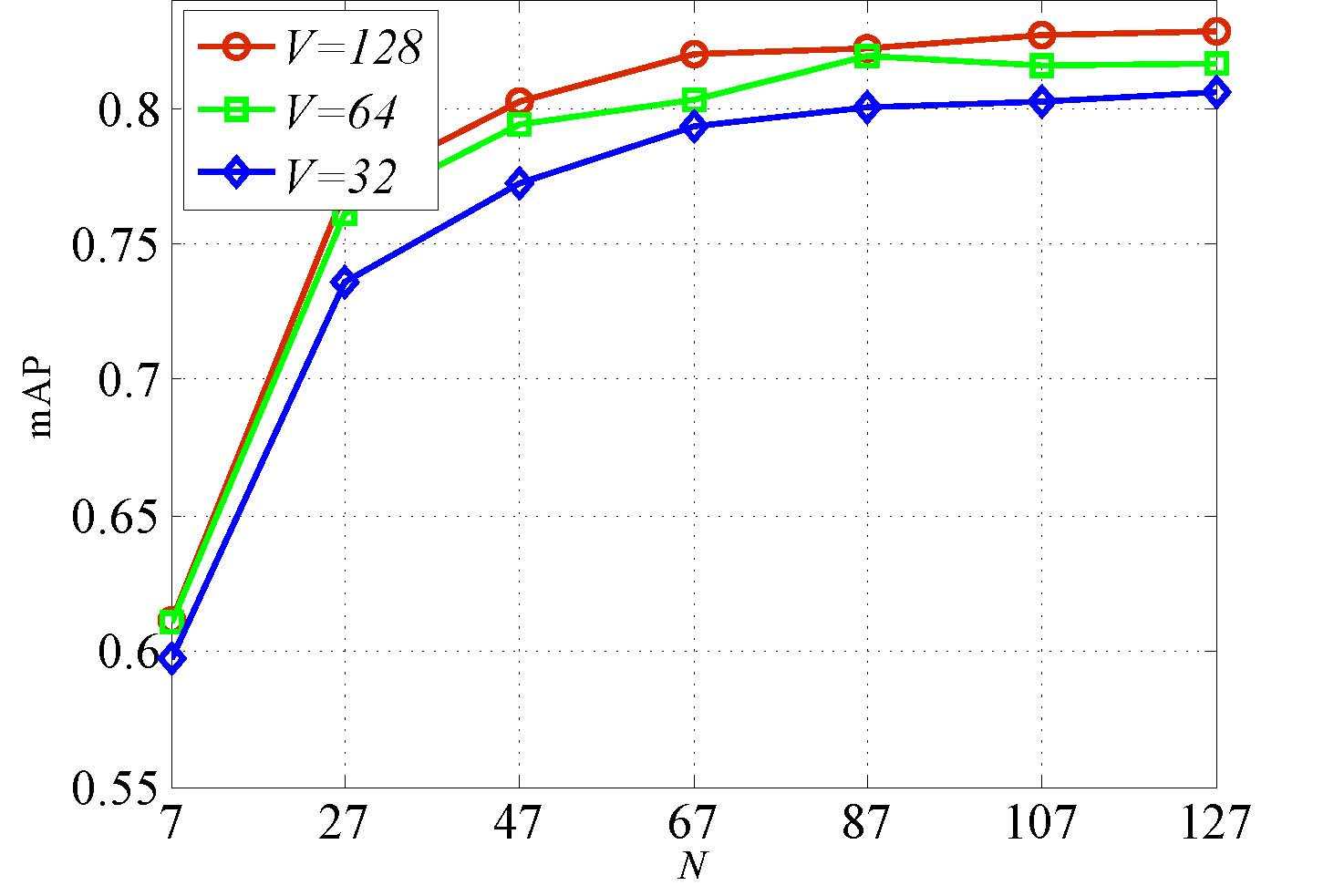}}
\end{center}
\caption{The illustration of the impact of the parameters in the proposed kernelized convolutional neural network (KCNN) algorithm on Holidays dataset. (a), (b), (c) are the results when image patch $x$ is not rotated. (d), (e), (f) are the similar meanings but with $x$ rotated. $N$ is the number of object detected with BING detector \cite{BingObj2014}. $V$ is the number of Gaussian functions used in Fisher vector model \cite{FisherVector}. $D$ is the dimension of the CNN features \cite{jia2014caffe} after performing the PCA dimension reduction.}
\label{result_1}
\end{figure}

\renewcommand{\arraystretch}{1.1}

\begin{table}[t]
\centering
\caption{The performance of the proposed kernelized convolutional neural network (KCNN) algorithm on three benchmark datasets, namely Holidays \cite{Hamming:Embedding}, Oxford Building \cite{AKM}, and UKBench \cite{vocabuary:tree}. $D=128$, $N=127$ are used here.}
\tabcolsep 0.03in
\begin{tabular}{|c|c|c|c|c|c|} \hline
\multirow{3}{*}{Dataset} &\multirow{3}{*}{CNN} &  \multicolumn{4}{|c|}{KCNN} \\ \cline{3-6}

 & & \multicolumn{2}{|c|}{Non Rotated $x$} & \multicolumn{2}{|c|}{Rotated $x$} \\ \cline{3-6}

 & & {$V$=64} & {$V$=128} & {$V$=64} & {$V$=128}\\ \hline

{Holidays} & \multirow{2}{*}{0.68} & {0.793} & {0.801} & {0.823} & {\textbf{0.829}} \\
 (mAP) &  & (+17.7\%) &  (+17.8\%) & (+21\%)  & (+\textbf{21.9\%})  \\ \hline

{UKBench} & \multirow{2}{*}{3.41} & {3.46} & {3.51} & {3.72} & {\textbf{3.74}} \\
 (top-4) &  & (+1.5\%) & (+2.9\%) & (+9.1\%) & (+\textbf{9.7\%})  \\ \hline

{Oxford} & \multirow{2}{*}{0.38} & {0.48} & {\textbf{0.51}} & {0.42} & {0.45} \\
 (mAP) &  & (+26.3\%) & (+\textbf{34.2\%}) & (+10.5\%)  & (+18.4\%) \\ \hline

\end{tabular}
\label{results_2}
\end{table}

In this section, we study the impact of parameters.
There are three parameters in our algorithm. The first one is the number of image patches $x$ detected by BING detector \cite{BingObj2014}, which can be denoted by $N$. The second one is the dimension of vectorial representation of image patch, $\gamma(x)$. We adopt the CNN model to construct the vectorial representation of $x$ resulting in a 4096-D $\gamma(x)$ \cite{jia2014caffe}. For convenience without loss of generality, we perform the principle components analysis (PCA) to reduce the 4096-D $\gamma(x)$ to $D$ dimension. The last parameter is the visual vocabulary size used in Fisher vector \cite{FisherVector} corresponding to the $\beta(\cdot)$ in Eq. \ref{KCNN_e}, which can be denoted by $V$.

\begin{figure}
\begin{center}
 \subfigure[]{
 \label{ukbench_samples_f}
 \includegraphics[width=0.45 \linewidth]{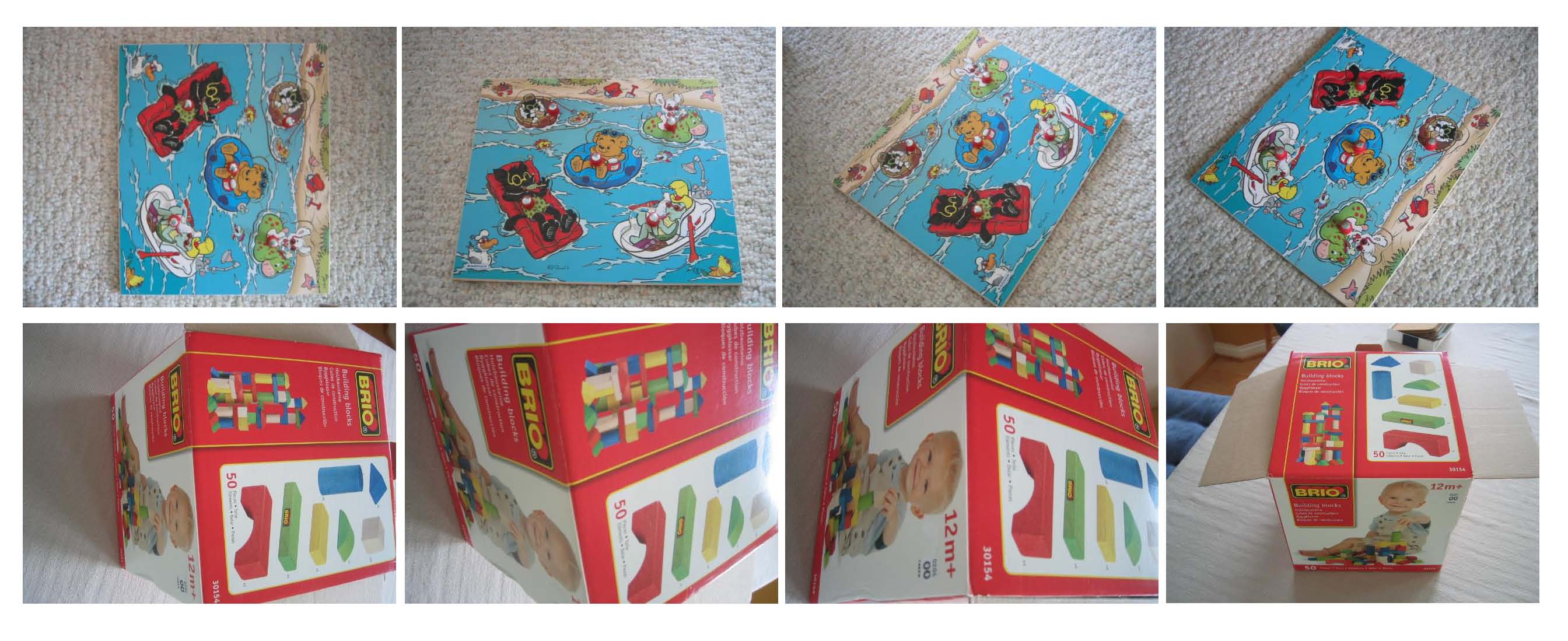}}
  \subfigure[]{
  \label{oxford_samples_f}
 \includegraphics[width=0.45 \linewidth]{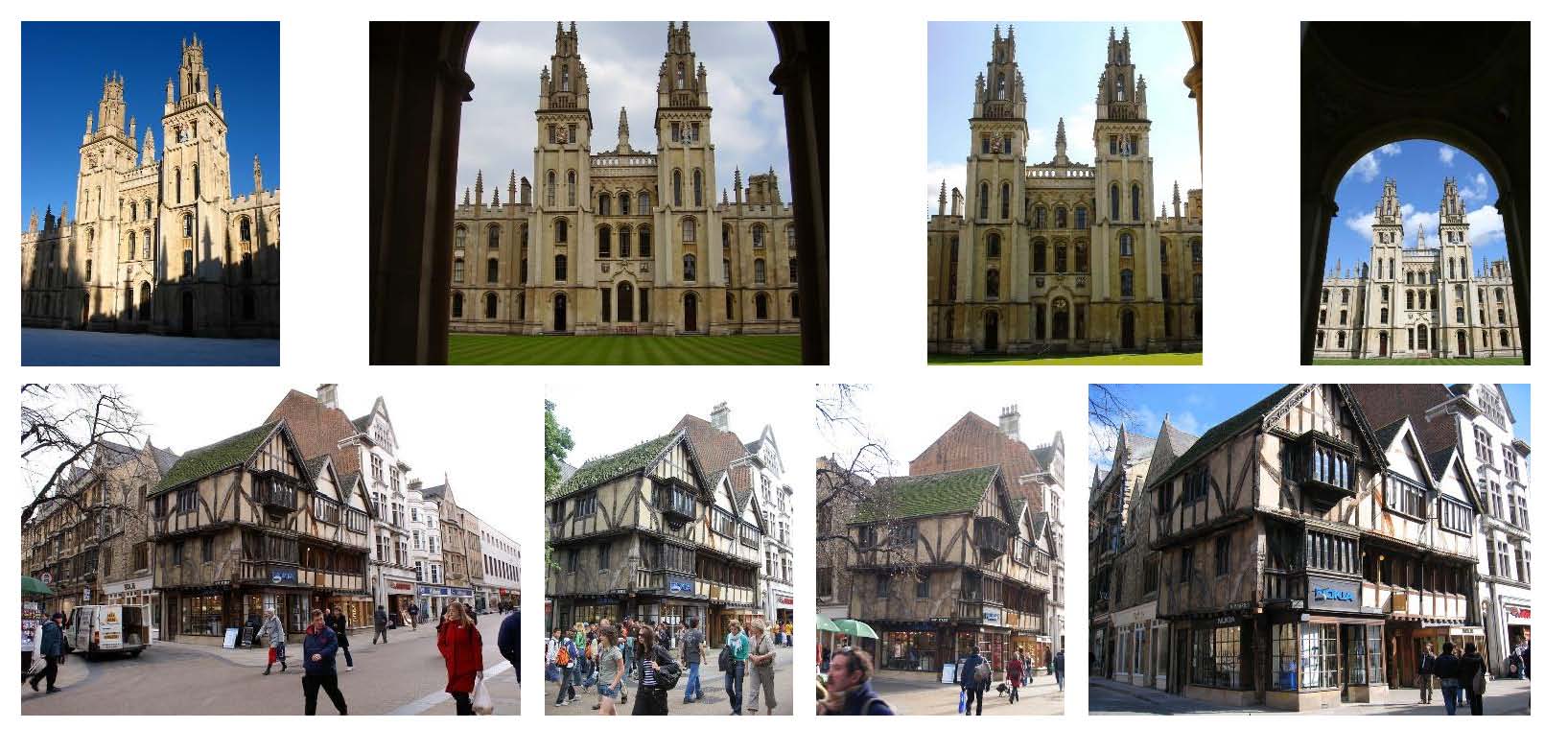}}
\end{center}
\caption{The examples of the transformations between images. (a) UKBench dataset (b) Oxford Building dataset}
\end{figure}

The results are demonstrated in Fig. \ref{result_1}. It can be seen that better accuracy can be obtained when more patches (larger $N$) are used. Similarly with larger $D$ and $V$, we can obtain higher accuracy. However the impacts of $D$ and $V$ are minor than $N$. In Table \ref{results_2}, we demonstrate the performance of the proposed KCNN algorithm on Holidays, UKBench, and Oxford Building datasets. We can see that it is benefical to perform rotation operation to image patch $x$ for Holidays and UKBench dataset. Especially for the UKBench dataset, the accuracy for the CNN feature is 3.41 and is improved to 3.51 (+2.9\%) with our KCNN algorithm without rotating $x$. After performing the rotation to $x$, the accuracy is improved from 3.51 to 3.74 (+6.6\%). This is because there are many rotation transformations in UKBench dataset, as shown in Fig. \ref{ukbench_samples_f}. However, the rotation operation to image patch $x$ is harmful on Oxford Building dataset for our KCNN algorithm. Similar result has also been observed when SIFT features are used to perform retrieval on this dataset \cite{perd2009efficient} \cite{jegou2012aggregating}. That is, in the construction of SIFT descriptor, better retrieval performance is obtained with the orientation selected as the gravity orientation instead of the traditional dominant gradient orientation \cite{SIFT} \cite{jegou2012aggregating}, since there is very few rotation transformations for the building images, as demonstrated in Fig. \ref{oxford_samples_f}.

\begin{figure}[t]
\begin{center}
 \subfigure[]{
 \includegraphics[width=0.6 \linewidth]{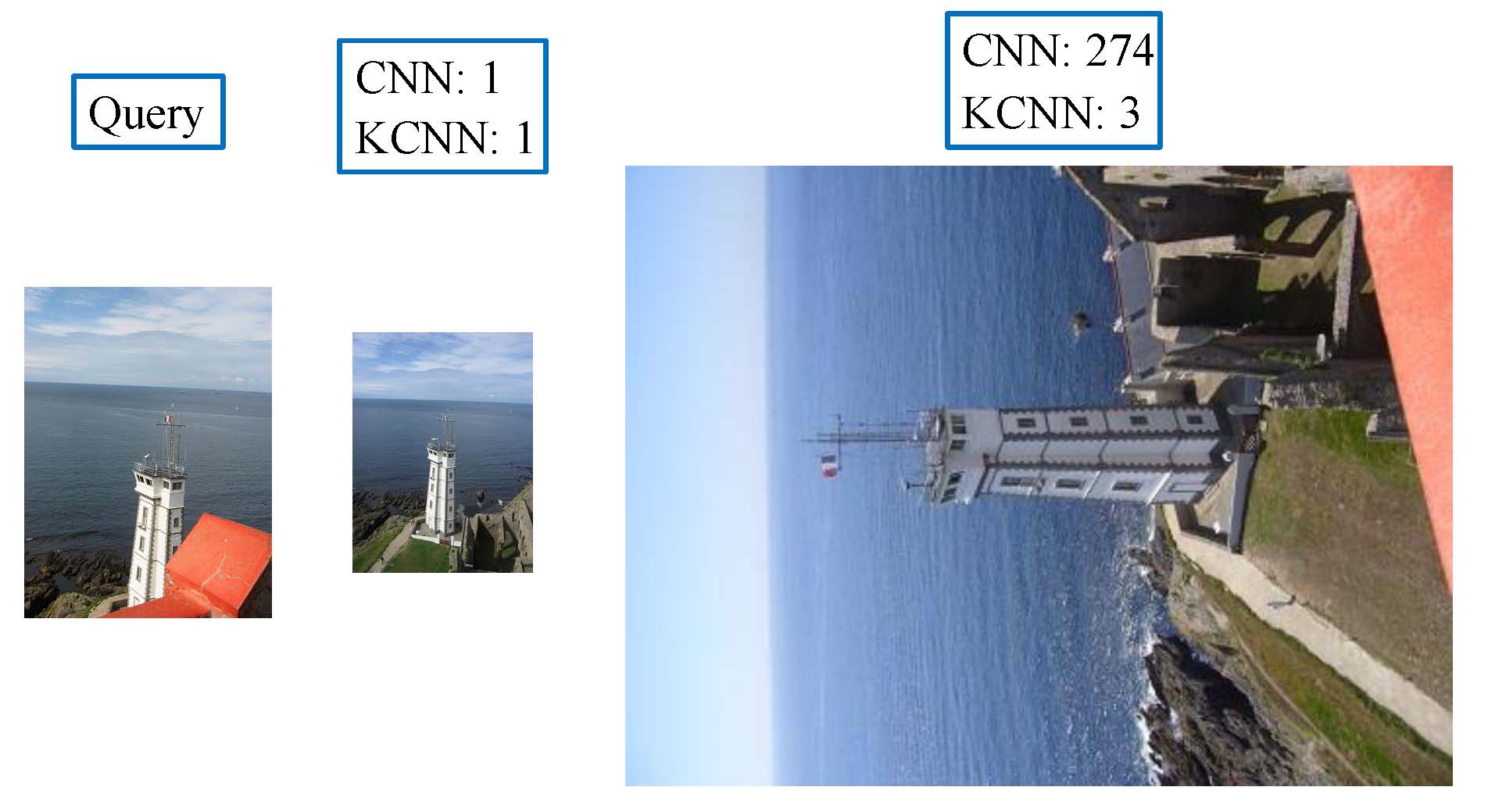}
 \label{h_e_1}
 }
 \subfigure[]{
 \includegraphics[width=0.6 \linewidth]{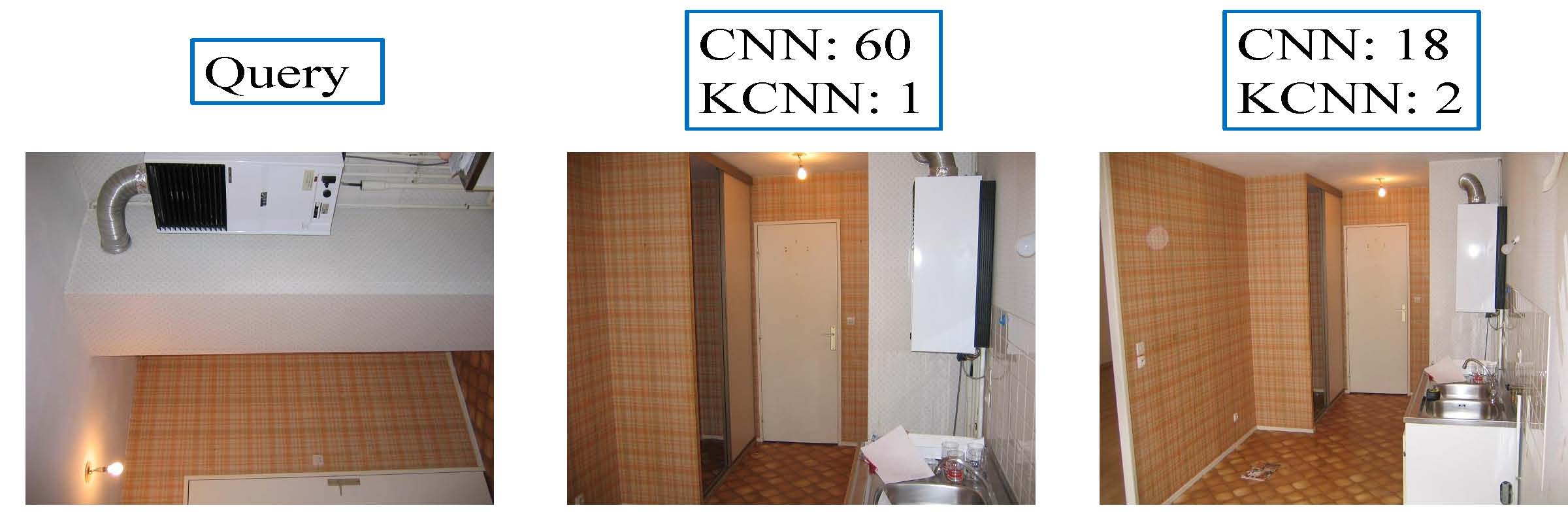}
  \label{h_e_2}
 }
 \subfigure[]{
 \includegraphics[width=0.6 \linewidth]{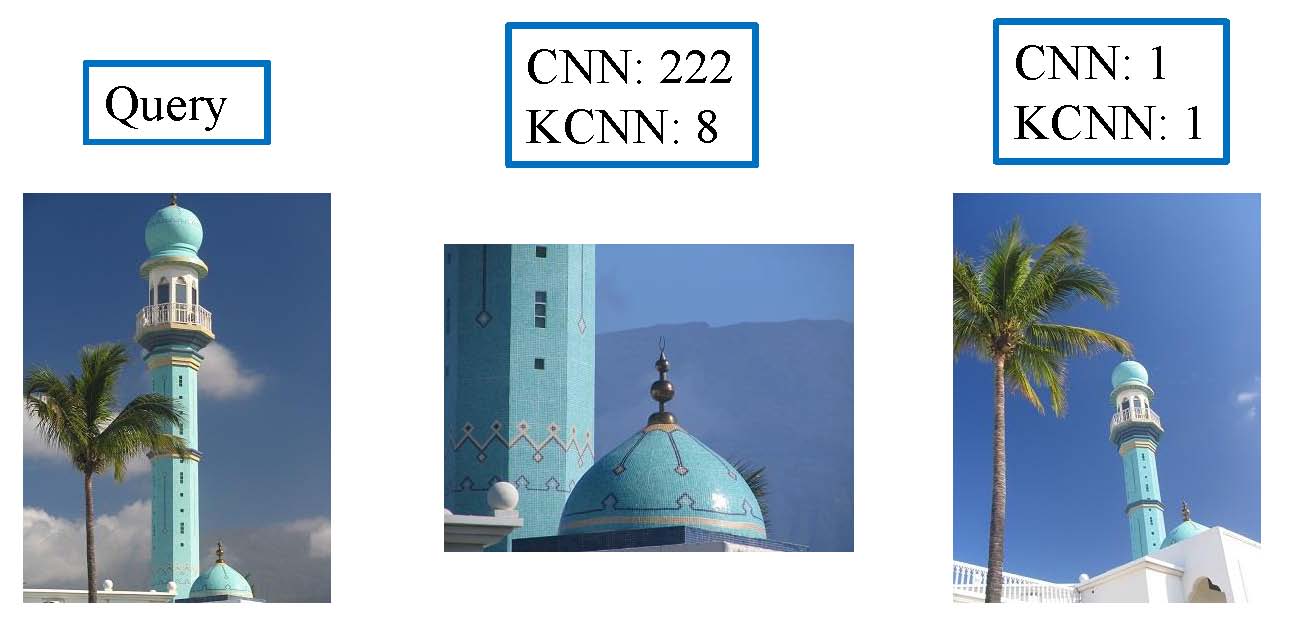}
  \label{h_e_3}
 }
\end{center}
\caption{Some examples of search results on Holidays dataset. Their rank numbers are also given in the blue boxes with CNN feature and KCNN feature.}
\label{holidays_examples}
\end{figure}

To further demonstrate the performance of the proposed kernelized convolutional neural network (KCNN) algorithm, we show the Average Precision (AP) of each query of Oxford Building dataset in Table \ref{results_3}. It can be seen that the proposed KCNN algorithm can get better retrieval application than the original convolutional neural network (CNN) algorithm for most queries. There are 38 queries out of total 55 queries (69.1\%) whose retrieval performance have been improved. The highest improvement comes from the query ``ashmolean\_2" whose retrieval performance is improved by 355.3\% from 0.0987 to 0.4495. Some examples on Holidays dataset are shown in Fig. \ref{holidays_examples}, in which we give their rank number with CNN representation and our KCNN representation. From Fig. \ref {h_e_1} and Fig. \ref{h_e_2}, it can be seen that our KCNN well addresses the sensitivity of rotation transformation which may fail the global CNN feature. From the first result of Fig. \ref{h_e_1} and the second result of Fig. \ref{h_e_3}, it can be seen that global CNN can also tolerate the slightly scaling transformation while our KCNN can do much better as shown in the first result of Fig. \ref{h_e_3}.


\begin{table*}[t]
\centering
\caption{The detailed results of each query on Oxford Building dataset.}
\tabcolsep 0.01in
\begin{tabular}{|c|c|c|c|c|c|c|c|c|c|c|c|c|c|c|c|} \hline
{Average} & \multicolumn{5}{|c|}{all\_souls} & \multicolumn{5}{|c|}{ashmolean} & \multicolumn{5}{|c|}{balliol} \\ \cline{2-16}

{Precision} & 1 & 2 & 3 & 4 & 5 & 1 & 2 & 3 & 4 & 5 & 1 & 2 & 3 & 4 & 5  \\ \hline

{CNN} & 0.113 & 0.21 & 0.274 & 0.546 & 0.156 & 0.366 & 0.099 & 0.068 & 0.229 & 0.208 & 0.134 &
0.121 & 0.206 & 0.322 & 0.302   \\ \hline

{KCNN} & 0.32 & 0.416 & 0.447 & 0.748 & 0.491 & 0.733 & 0.449 & 0.293 & 0.717 & 0.272 & 0.581 & 0.414 & 0.388 & 0.603 & 0.447  \\ \hline

{Improved} & +184\% & +98\% & +63\% & +37\% & +215\% & +100\% & +355\% & +334\% & +213\% & +31\% & +333\% & +242\% & +89\% & +87\% & +48\%  \\ \hline

\cline{1-16}
\cline{1-16}
\cline{1-16}
\cline{1-16}
{Average} & \multicolumn{5}{|c|}{bodleian} & \multicolumn{5}{|c|}{christ\_church} & \multicolumn{5}{|c|}{cornmarket}  \\ \cline{2-16}
{Precision} & 1 & 2 & 3 & 4 & 5 & 1 & 2 & 3 & 4 & 5 & 1 & 2 & 3 & 4 & 5   \\ \hline

{CNN}& 0.207 & 0.262 & 0.421 & 0.463 & 0.431 & 0.442 & 0.487 & 0.363 & 0.213 & 0.144 & 0.594 & 0.223 & 0.133 & 0.139 & 0.559  \\ \hline

{KCNN}& 0.502 & 0.592 & 0.535 & 0.598 & 0.63 & 0.472 & 0.56 & 0.526 & 0.399 & 0.134 & 0.426 & 0.4 & 0.307 & 0.515 & 0.578  \\ \hline
{Improved} & +142\% & +126\% & +27\% & +29\% & +46\% &+6.8\% & +15\% & +45\% & +87\% & -6.8\%  & -28\% &+80\% &+130\% &+271\% &+3.5\%  \\
\cline{1-16}
\cline{1-16}
\cline{1-16}
\cline{1-16}

{Average} & \multicolumn{5}{|c|}{hertford} & \multicolumn{5}{|c|}{keble} & \multicolumn{5}{|c|}{magdalen}  \\ \cline{2-16}

{Precision} & 1 & 2 & 3 & 4 & 5 & 1 & 2 & 3 & 4 & 5 & 1 & 2 & 3 & 4 & 5   \\ \cline{1-16}

{CNN} & 0.6 & 0.617 & 0.588 & 0.636 & 0.64 & 0.383 & 0.482 & 0.551 & 0.209 & 0.336  & 0.104 & 0.114 & 0.087 & 0.143 & 0.081  \\ \cline{1-16}

{KCNN} & 0.605 & 0.604 & 0.586 & 0.609 & 0.471 & 0.696 & 0.747 & 0.638 & 0.547 & 0.226 & 0.09 & 0.08 & 0.078 & 0.1 & 0.08   \\ \cline{1-16}
{Improved} &+0.8\% &-2.2\%&-0.4\%&-4.3\%&-27\% &+82\%&+55\%&+16\%&+162\%&-33\% &-13\%&-30\%&-11\%&-30\%&-2\%\\ \cline{1-16}

\cline{1-16}
\cline{1-16}
\cline{1-16}
\cline{1-16}

{Average}  & \multicolumn{5}{|c|}{pitt\_rivers} & \multicolumn{5}{|c|}{radcliffe\_camera} & \multicolumn{5}{c}{} \\ \cline{2-11}
{Precision} & 1 & 2 & 3 & 4 & 5 & 1 & 2 & 3 & 4 & 5    \\ \cline{1-11}

{CNN} & 0.563 & 0.678 & 0.45 & 0.472 & 0.519 & 0.876 & 0.788 & 0.89 & 0.889 & 0.889 \\ \cline{1-11}
{KCNN} & 0.651 & 0.846 & 0.662 & 0.381 & 0.621 & 0.743 & 0.842 & 0.842 & 0.876 & 0.709  \\ \cline{1-11}
{Improved} &+15.6\%&+24.8\%&+47\%&-19.5\%&+19.7\%&-15.2\%&+6.9\%&-5.3\%&-1.5\%&-20.2\%  \\ \cline{1-11}

\end{tabular}
\label{results_3}
\end{table*}

\subsection{Comparisons}
In this section, we give some comparisons with the results reported in other research works.
As shown in Table \ref{comparisons}, it can be seen that the proposed KCNN method obtains best result on both Holidays and UKBench datasets. However on Oxford Building dataset SIFT \cite{SIFT} based methods can get better result namely \cite{FisherVector} \cite{all_vlad} \cite{jegou:hal-00977321}. The reason is that the Oxford Building dataset consists of building images and the retrieval on this dataset is more like a fine-grained problem \cite{nilsback2008automated}. On the other hand deep convolutional neural network is designed to tackle the generic classification problem \cite{image-net, krizhevsky2012imagenet} and fine-tune is usually required for the fine-grained vision tasks.

There also exists some works on performing image search with CNN. However our work has substantial difference with them. Comparing with \cite{razavian2014cnn}, our goal is totally different. Our goal is to construct a vectorial representation for an image while \cite{razavian2014cnn} use the Spatial Search that is not a vectorial image representation. The spatial search means extensively search all the sub-patches extracted on the grids at several levels. The search complexity will be $O(N^2)$ where $N$ means the number of extracted sub-patches. Comparing with \cite{gong2014multi}, on Holidays, we get 0.829 mAP while \cite{gong2014multi} gets 0.802 mAP. Besides the higher accuracy, we also address the rotation transformation while \cite{gong2014multi} not. Comparing with \cite{babenko2014neural}, they focus on construct compressed codes of image representation with the retrained regular CNN while we focus on addressing the object transformations in the vectorial representation of complex images without retraining.

\begin{table}[t]
\centering
\caption{The performance comparisons with other reported research works based on global image representation.}
\tabcolsep 0.05in
\begin{tabular}{|c|c|c|c|c|c|} \hline
{Dataset} & \cite{FisherVector} & \cite{all_vlad} & \cite{jegou:hal-00977321} & {CNN} &  {KCNN} \\ \hline

{Holidays} & \multirow{2}{*}{0.735} & \multirow{2}{*}{0.646} & \multirow{2}{*}{0.771} & \multirow{2}{*}{0.68} & \multirow{2}{*}{0.829}   \\
 (mAP) &  & & & &\\ \hline

{UKBench} & \multirow{2}{*}{3.50} & \multirow{2}{*}{N/A} & \multirow{2}{*}{3.53} & \multirow{2}{*}{3.41} & \multirow{2}{*}{3.74}   \\
 (top-4) &  &  &&& \\ \hline

{Oxford Building} & \multirow{2}{*}{N/A} & \multirow{2}{*}{0.555} & \multirow{2}{*}{0.676} & \multirow{2}{*}{0.38} & \multirow{2}{*}{0.51}  \\
 (mAP) &  & && & \\ \hline

\end{tabular}
\label{comparisons}
\end{table}

\section{Conclusion} \label{conclusion}

In this paper, we have analyzed the sensitivity of the global CNN feature to the geometric transformations of image such as translation, scaling, and rotation. Based on our analysis, inspired by the well-studied local feature based image representation methods, we proposed our kernelized convolutional network (KCNN) algorithm to describe the content of complex images. With our KCNN method, we can obtain a more robust vectorial representation. Besides the CNN structure implemented in Caffe library, there are also some other emerging CNN structures \cite{simonyan2014very, iandola2014densenet}. In the future, we would like to investigate the potential of these different CNN models on image retrieval and investigate the performance of our KCNN model integrated with these CNN models.

{\small
\bibliographystyle{ieee}
\bibliography{sigproc_full_2}
}

\end{document}